\documentclass[final,12pt]{clear2025} 

\title[Optimizing Multi-Scale Representations to Detect Effect Heterogeneity]{Optimizing Multi-Scale Representations to Detect Effect Heterogeneity Using Earth Observation and Computer Vision: Applications to Two Anti-Poverty RCTs}
\usepackage{times}

\usepackage{tikz}
\usetikzlibrary{arrows.meta, positioning}

\usepackage[utf8]{inputenc} 
\usepackage[T1]{fontenc}    
\usepackage{url}            
\usepackage{booktabs}       
\usepackage{amsfonts}       
\usepackage{nicefrac}       
\usepackage[protrusion=true, expansion=true]{microtype}
\usepackage{xcolor}         
\usepackage[format=plain,
            labelfont={bf,it},
            textfont=it]{caption}

\newcommand\PerulcliplrsicdlTRUEMeanDiff{0.00}
\newcommand\PerulcliplrsicdlTRUEMeanDifflse{0.06}
\newcommand\PerulcliplrsicdlTRUEMeanDifflpc{0.68}
\newcommand\PerulcliplrsicdlTRUEMeanDifflpclse{0.07}
\newcommand\UgandalcliplrsicdlTRUEMeanDiff{0.19}
\newcommand\UgandalcliplrsicdlTRUEMeanDifflse{0.08}
\newcommand\UgandalcliplrsicdlTRUEMeanDifflpc{1.35}
\newcommand\UgandalcliplrsicdlTRUEMeanDifflpclse{0.10}
\newcommand\PerulcliplrsicdlFALSEMeanDiff{0.00}
\newcommand\PerulcliplrsicdlFALSEMeanDifflse{0.08}
\newcommand\PerulcliplrsicdlFALSEMeanDifflpc{0.68}
\newcommand\PerulcliplrsicdlFALSEMeanDifflpclse{0.09}
\newcommand\UgandalcliplrsicdlFALSEMeanDiff{0.41}
\newcommand\UgandalcliplrsicdlFALSEMeanDifflse{0.09}
\newcommand\UgandalcliplrsicdlFALSEMeanDifflpc{0.95}
\newcommand\UgandalcliplrsicdlFALSEMeanDifflpclse{0.10}

\newcommand\AveDisplaced{2.24}
\newcommand\AveNonDisplaced{2.13}

\usepackage{placeins}

\usepackage{booktabs}
\usepackage{graphicx}
\usepackage{float}
\usepackage{lmodern}
\usepackage{latexsym,multirow}
\usepackage{amssymb,amsmath,bm}
\usepackage{bigints}
\usepackage{longtable}
\usepackage{bbm}
\usepackage[color,all,import,arrow]{xy}
\usepackage{array}
\usepackage{enumerate}
\usepackage{verbatim}
\usepackage{mathtools}
\usepackage{multicol}
\usepackage[utf8]{inputenc}

\usepackage{dcolumn}
\newcolumntype{.}{D{.}{.}{-1}}
\newcolumntype{d}[1]{D{.}{.}{#1}}

\graphicspath{{./}}

\usepackage[T1]{fontenc}
\usepackage{tikz}
\usepackage{qtree}
\usepackage{comment}
\usetikzlibrary{arrows}
\usetikzlibrary{positioning}
\usetikzlibrary{matrix}
\usetikzlibrary{bayesnet}
\usetikzlibrary{decorations.pathreplacing,calligraphy}

\usepackage{xcolor}

\theoremheaderfont{\scshape}

\usepackage{rotating}

\usepackage{arydshln}

\usepackage[compact]{titlesec}

\usepackage{setspace}
\usepackage[bottom]{footmisc}
\allowdisplaybreaks

\newcommand{\E}{\mathbb{E}}

\newcommand{\bx}{\mathbf{x}}

\newcommand{\bM}{\mathbf{M}}
\newcommand{\bbm}{\mathbf{m}}

\newcommand{\bW}{\mathbf{W}}

\newcommand{\bZ}{\mathbf{Z}}
\newcommand{\bY}{\mathbf{Y}}

\newcommand{\bphi}{\boldsymbol{\phi}}

\clearauthor{%
 \Name{Fucheng Warren Zhu} \Email{\href{mailto:wzhu@college.harvard.edu}{\texttt{wzhu@college.harvard.edu}}}\\
 \addr Department of Statistics, Harvard University \\
 Department of Computer Science, Harvard University
 \AND
 \Name{Connor T.\, Jerzak} \Email{\href{mailto:connor.jerzak@austin.utexas.edu}{\texttt{connor.jerzak@austin.utexas.edu}}}\\
 \addr Department of Government, University of Texas at Austin%
 \AND
 \Name{Adel Daoud} \Email{\href{mailto:adel.daoud@liu.se}{\texttt{adel.daoud@liu.se}}}\\
 \addr Institute for Analytical Sociology, Link\"{o}ping University 
 \\ Chalmers University of Technology
}

\begin{document}

\maketitle

\begin{abstract}%
  \noindent Earth Observation (EO) data are increasingly used in policy analysis by enabling granular estimation of conditional average treatment effects (CATE). However, a challenge in EO-based causal inference is determining the scale of the input satellite imagery---balancing the trade-off between capturing fine-grained individual heterogeneity in smaller images and broader contextual information in larger ones. This paper introduces Multi-Scale Representation Concatenation, a set of composable procedures that transform arbitrary single-scale EO-based CATE estimation algorithms into multi-scale ones. We benchmark the performance of Multi-Scale Representation Concatenation on a CATE estimation pipeline that combines Vision Transformer (ViT) models (which encode images) with Causal Forests (CFs) to obtain CATE estimates from those encodings. We first perform simulation studies where the causal mechanism is known, showing that our multi-scale approach captures information relevant to effect heterogeneity that single-scale ViT models fail to capture as measured by $R^2$. We then apply the multi-scale method to two randomized controlled trials (RCTs) conducted in Peru and Uganda using Landsat satellite imagery. As we do not have access to ground truth CATEs in the RCT analysis, the Rank Average Treatment Effect Ratio (RATE Ratio) measure is employed to assess performance. Results indicate that Multi-Scale Representation Concatenation improves the performance of deep learning models in EO-based CATE estimation without the complexity of designing new multi-scale architectures for a specific use case. The application of Multi-Scale Representation Concatenation could have meaningful policy benefits---e.g., potentially increasing the impact of poverty alleviation programs without additional resource expenditure.

\end{abstract}


\section{Introduction}

Earth Observation (EO) data play an increasingly important role in policy analysis by providing researchers with contextual information to estimate treatment effects at a more granular level, revealing characteristics of environmental conditions, land use patterns, economic development, and climate variables \citep{anderson2017earth,daoud_using_2023-1,burkeUsingSatelliteImagery2021,pettersson2023time,kino_scoping_2021}. A growing body of work therefore focuses on estimating household or neighborhood-specific Conditional Average Treatment Effects (CATE) \citep{sakamoto2024scopingreviewearthobservation, jerzak_image-based_2023, serdavaa_satellite_2023, giannarakis_understanding_2023, go_use_2022,daoudImpactAusterityChildren2024,shibaUncoveringHeterogeneousAssociations2022a}. 

An important application of EO-based policy analysis lies in anti-poverty Randomized Controlled Trials (RCTs), where the outcome $\bY$ is the household income for the unit that the satellite imagery is centered on, and the treatment $\bW$ is cash transfer to the household. The researcher's goal is to estimate effects conditioned on available satellite imagery $\bM$. We denote this estimate for a given image, $\hat{\tau}(\bbm)$. In order to do so, we encode $\bM$ into a sufficiently low-dimensional representation, $\phi$, that preserves maximal information about $\bZ$, the set of unobserved background features that are relevant to treatment (e.g., the village geography/baseline wealth). See the white nodes of Figure \ref{fig:DAG} for illustration.

\begin{figure}[htb]
    \centering
\includegraphics[width=0.2\textwidth,height=0.1\textheight]{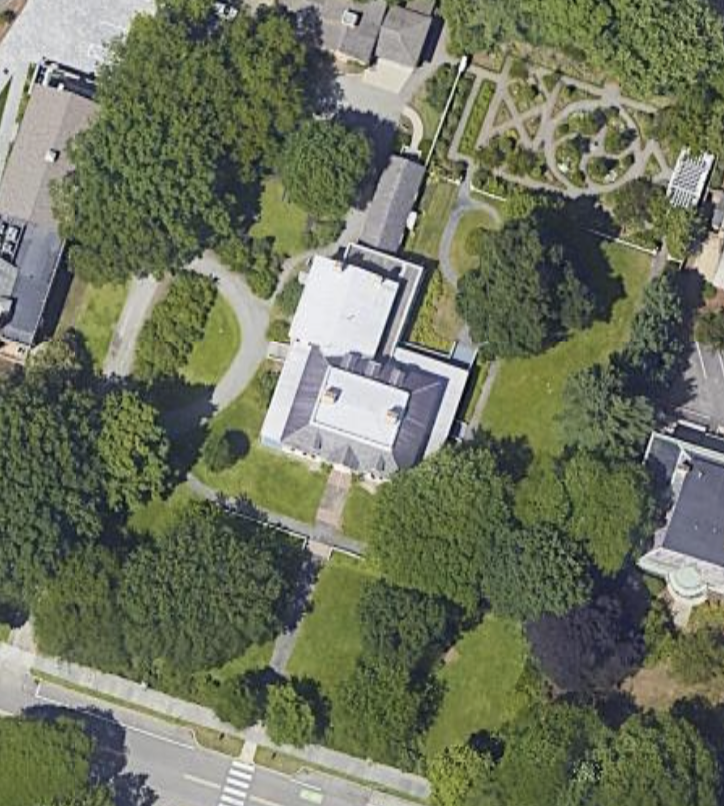}  
\includegraphics[width=0.2\textwidth,height=0.1\textheight]{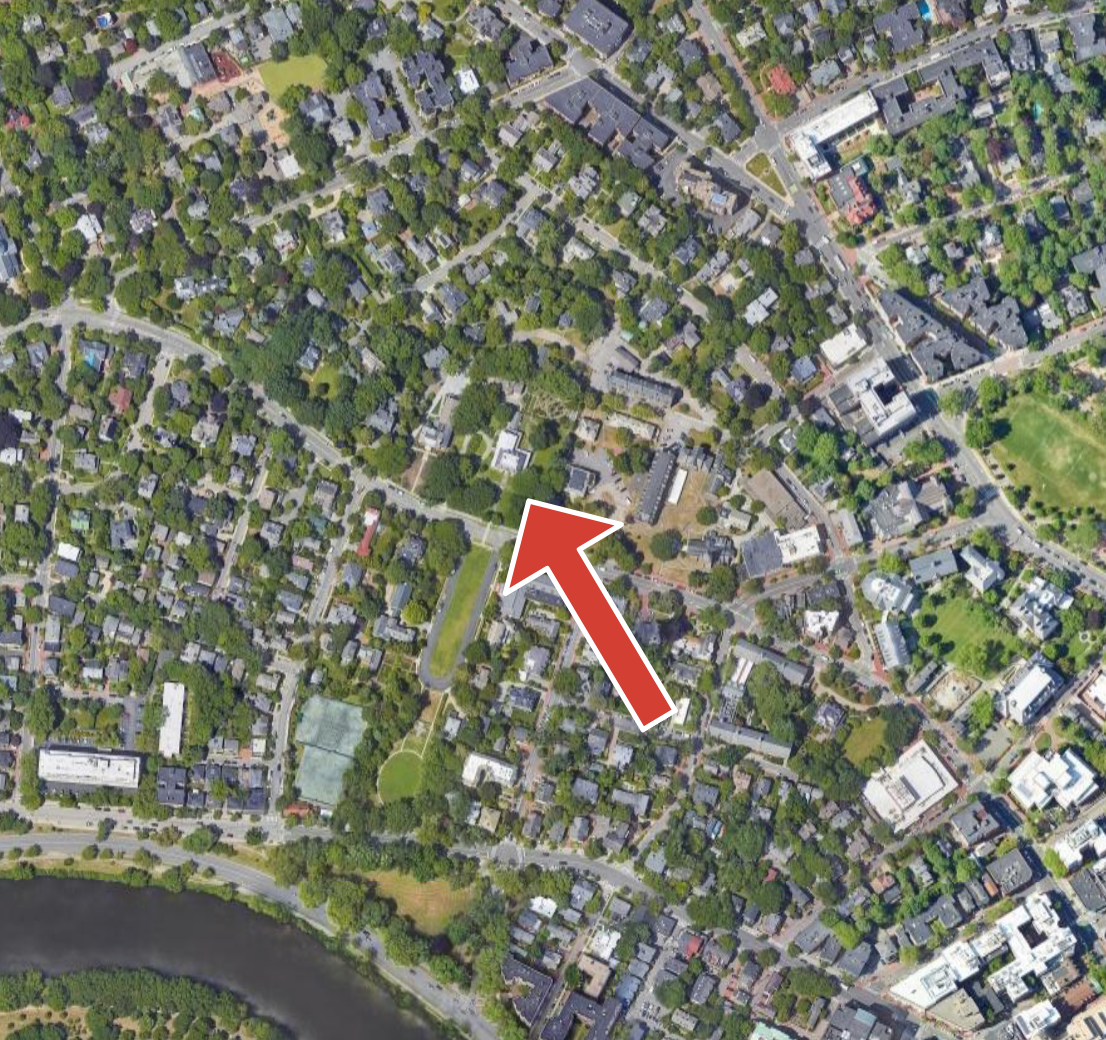} 
\caption{The Washington-Longfellow National Historic Site {\scshape (Left)}, with context {\scshape (Right)}. Using sentences in the RSICD dataset, CLIP's nearest-neighbor text embedding to the images are ``the square with some halls and lawns is in a rectangular region which (sic.) is in the middle of the forest'' and ``here is a zone of apartment buildings between a river and a road' respectively, showing that the model extracts separate, novel information from the two images.
}
    \label{fig:CloseFarIll}
\end{figure}

A key characteristic of EO-based images in the context of effect heterogeneity estimation is the inherent multi-scale nature of the relevant background  features $\bZ$ for observational units, a phenomenon termed ``multi-scale dynamics'' \citep{xiong_confounder-free_2022, reed2023scalemaescaleawaremaskedautoencoder}. Figure \ref{fig:CloseFarIll} provides some intuition for such multi-scale dynamics in EO-based inference, where information relevant to heterogeneity is encoded in both the local area around a household and the broader neighborhood around which the household is situated. 



Along with Figure \ref{fig:CloseFarIll}, we provide a visualization in Figure \ref{fig:DAG} that provides a more formal illustration of our setting. Parts highlighted in blue refer to multi-scaled components of the DAG that were previously not considered in effect heterogeneity estimation methodologies.

\begin{figure}[htbp]
\centering
\begin{tikzpicture}[
  >=Stealth,
  node distance=2.2cm,
  baseNode/.style={
    scale=0.55,
    font=\small, 
    draw, 
    rectangle, 
    rounded corners, 
    align=center
  },
  newNode/.style={
    baseNode,
    draw=blue!60,
    fill=blue!10
  },
  newEdge/.style={
    ->,
    color=blue!60
  },
  every edge/.append style={>=Stealth}
]

\node (T) [baseNode] {%
  \(W_{i}\)\\[-1pt]
  (Treatment, \\ observed)
};

\node (Z1) [baseNode, dashed, below=0.6cm of T] {%
  \(\mathbf{Z}_{i, s_1}\)\\[-1pt]
  (Unobserved factors, \\ contextual-scale)
};

\node (Y) [baseNode, right=3cm of T] {%
  \(Y_i\)\\[-1pt]
  (Observed \\ outcome)
};

\node (M1) [baseNode, right=2cm of Z1] {%
  \(\mathbf{M}_{i, s_1}\)\\[-1pt]
  (Larger-scale\\ imagery)
};

\node (phi) [baseNode, right=3cm of M1] {%
  \(\bphi_{i}\)\\[-1pt]
  (Concatenated\\ representations)
};

\node (tau) [baseNode, above=0.6cm of phi] {%
  \(\hat{\tau}_i\)\\[-1pt]
  (CATE estimate)
};

\node (Z2) [newNode, dashed, below=0.6cm of Z1] {%
  \(\mathbf{Z}_{i, s_2}\)\\[-1pt]
  (Unobserved factors, \\ unit-scale)
};

\node (M2) [newNode, right=1.9cm of Z2] {%
  \(\mathbf{M}_{i, s_2}\)\\[-1pt]
  (Smaller-scale\\ imagery)
};

\node (phi1) [baseNode, right=0.8cm of M1, yshift=0.4cm] {%
  \(\bphi_{i, s_1}\)\\[-1pt]
  (Larger-\\ scale \\ reps)
};
\node (phi2) [newNode, right=0.7cm of M2, yshift=0.0cm] {%
  \(\bphi_{i, s_2}\)\\[-1pt]
  (Smaller-\\scale \\ reps)
};

\draw[->] (T) -- (Y);
\draw[->] (Z1) -- (Y);
\draw[->] (Z1) -- (M1);
\draw[->] (phi) -- (tau);
\draw[->, out=15, in=165, looseness=1.1] (T.north) to (tau.north);
\draw[->] (Y) -- (tau);

\draw[newEdge] (Z2.north east) -- (Y);
\draw[newEdge] (Z2) -- (M2);

\draw[->] (M1) -- (phi1);
\draw[newEdge] (M2) -- (phi2);

\draw[->] (phi1) -- (phi);
\draw[newEdge] (phi2) -- (phi);

\end{tikzpicture}

\caption{A DAG showing the relationship between the various components involved in CATE estimation. The $i$ subscript is the observation unit index; the subscript $s$ ($s_1$ or $s_2$) is the scale of the variable. The blue components are the components of the figure that reflect multi-scale dynamics. Multi-Scale Representation Concatenation allows algorithms that previously were only a function of $\bM_{i,s}$ for a single $s$ to utilize both $\bM_{i,s_1}$ and $\bM_{i,s_2}$. Note that this diagram could be extended to additional scales. Image resolution is assumed fixed.}

\label{fig:DAG}
\end{figure}

Our contribution is to propose Multi-Scale Representation Concatenation, a procedure that transforms single-scale EO-based CATE estimation algorithms into multi-scale ones by concatenating the image representation tensors. We benchmark using a single-scale CATE estimation algorithm previously used in anti-poverty RCTs \citep{jerzak_image-based_2023}. A remote-sensing fine-tuned image model, CLIP-RSICD, is used to generate lower-dimensional representations of single-scaled satellite imagery \citep{lu2017exploring,flax2021clip-rsicd-v2,radford2021CLIP}. A causal forest (CF) model then produces the final CATE estimates.\footnote{We also report results for SWIN and Clay model architectures, but use CLIP results as baseline due to the possibility of text-based interpretability analyses and the wide use of this model class.}

Beyond building on top of previous work in Causal Inference, there is a rich body of work investigating multi-scale phenomena in Computer Vision which we have drawn inspiration from to develop Multi-Scale Representation Concatenation. We survey them in \S\ref{s:relatedwork}. 

To quantify performance in real randomized controlled trials (RCTs) lacking ground truth CATE data, we use a measure of the degree of effect heterogeneity detected by a model, Rank Average Treatment Effect Ratio (RATE Ratio), as our metric \citep{yadlowsky2021evaluating}. We find evidence that applying Multi-Scale Representation Concatenation improves the RATE Ratio, and, thus, potentially improves accuracy of CATE estimates and policy targeting. 

As extensions, simulation studies are used to validate results from the RCT analysis due to the difficulty in evaluating model performance on RCT data where ground-truth CATE is unknown (\S\ref{s:Sim}); a scaling analysis demonstrates that Multi-Scale Representation Concatenation can significantly boost the performance of CATE estimation models in EO-based causal inference beyond just two scales (\S\ref{s:ScalingScale}); and an analysis where we assume the CATE analyst has limited prior information on the geographic distribution of household units finds that multi-scale inference improves the RATE metric under this weaker assumption (\S\ref{s:WeakPriorInformation}). These suggest promising further areas of exploration.




\section{Background and Contributions}

An enduring question in EO-based causal inference is how to best estimate $\tau(\bbm) \coloneqq \mathbb{E}[Y_i(1) - Y_i(0) \mid \bM_i = \bbm]$, i.e., the CATE for pre-treatment image array, $\bM_i$ \citep{athey2018impact, jerzak_image-based_2023}. In theory, one would provide all available covariates to an oracle function generating estimated $\tau(\bbm)$'s to obtain the most accurate causal estimates. For EO-based causal inference, this would ideally involve using the largest, highest-resolution satellite image available. However, in practice, this function is difficult to estimate due to the high dimensionality of $\bM_i$.

In this context, because there are well-understood methods to estimate CATEs from $\bW$, $\bY$, and pre-treatment covariate vectors \cite{chernozhukov2024appliedcausalinferencepowered}, a natural approach in image-based CATE estimation would be to introduce an encoder $\phi$ that maps high-dimensional imagery into a lower-dimensional representation. However, in an EO-based causal inference context, a pre-trained image encoder may struggle to capture both household-specific and neighborhood-level contextual information. In some cases, adding additional contextual information can even degrade performance if the model fails to distinguish relevant from non-relevant signals  \citep{damario2022dnndataefficiency}. Model fine-tuning could address this issue,  yet it remains difficult due to the small size of most RCTs.

Our main contribution therefore is to develop Multi-Scale Representation Concatenation, a family of procedures that transform any previous single-scale procedure into a multi-scale one through tensor concatenation. It is especially suitable in data-constrained settings (e.g. RCTs), as it requires minimal additional data requirements. The strong interpretability of the procedure also makes it simple to explain to relevant stakeholders. Further, it requires limited model experimentation: The optimal concatenation procedure can be found by systematically varying the image size and combining image representations from different scales. Multi-Scale Representation Concatenation thus transforms the problem of designing architectures specific to the multi-scale dynamics of the problem at hand to a less complex inference-time computational search \citep{snell2024scalingllmtesttimecompute}. We examine the improvements in model performance when incorporating Multi-Scale Representation Concatenation in the context of randomized controlled trials (RCTs) conducted in Peru and Uganda. 

\section{Methodology}

Let $i$ index the experimental units in the study. Each $i$ has a geolocation denoted by $\mathbf{x}_i \in \mathbb{R}^2$, representing spatial coordinates (e.g., latitude/longitude), and a binary treatment indicator $W_i$ encoding whether the unit received treatment. The observed outcome from the RCT for unit $i$ is $Y_i \in \mathbb{R}$. Let $Y_i(1)$ and $Y_i(0)$ denote potential outcomes under treatment and control. Identification will be performed assuming unconfoundedness and SUTVA \citep{imbens2016causal, chernozhukov2024appliedcausalinferencepowered}. 
Now define an image fetcher that generates an image from a coordinate ${\mathbf{x}_i} \in \mathbb{R}^2$ with a specified size $s \in \mathbb{N}_+$: $f_I: \mathbb{R}^2 \times \mathbb{N}_+ \to \mathcal{M}$, where $\mathcal{M}$ is the space of possible imagery: 
\vspace{-0.1cm}
\[
\bM_{i, s} = f_I(\mathbf{x}_i, s) \in \mathcal{M},
\]
\vspace{-0.1cm}
where $\bM_{i, s}$ is the image of size $s > 0$ centered at $\mathbf{x}_i$. Here, size refers to the width of the raw satellite imagery in terms of pixel number, with pixel resolution fixed.

Next, because of the high-dimensionality of $\bM_{i,s}$, we need to introduce a causal representation extraction function $\phi: \mathcal{M}^c \to \mathbb{R}^d$, which takes $c\in \mathbb{N}_+$ images as input and outputs a $d$-dimensional feature vector. The projection of the image to a lower dimensionality makes estimation tractable in finite samples. We develop a Multi-Scale Representation Concatenation procedure that constructs $\phi$ from base image encoders, $f_{\phi_{s_k}'}: \mathcal{M} \to \mathbb{R}^{d'}$, to easily adapt single-scale representation extraction functions into a multi-scale setting.

Although our algorithm is generalizable to arbitrary $c$, for concreteness, our exposition will be limited to the case where $c=2$, with $f_{\phi}: \mathcal{M} \times \mathcal{M} \to \mathbb{R}^{2d'}$:
\[
\boldsymbol{\phi}_{i, s_1, s_2} = f_{\phi}\left(\bM_{i, s_1},\, \bM_{i, s_2}\right) = \left( f_{\phi_{s_1}}'(\bM_{i, s_1}),\, f_{\phi_{s_2}}'(\bM_{i, s_2})\right),
\]
where $s_1$ and $s_2$ are two different image sizes. In this paper, $f_{\phi}$ is constructed through representation concatenation of the outputs of base image encoders $f_{\phi_{s_k}}'$ along with (optionally) a dimensionality reduction function $r$, so that $f_{\phi}\left(\bM_{i, s_1},\, \bM_{i, s_2}\right) = r((f_{\phi_{s_1}}'(\bM_{i, s_1}),f_{\phi_{s_2}}'(\bM_{i, s_2})))$. Here, $r: \mathbb{R}^{2d'} \to \mathbb{R}^l$, $l < 2d'$. 

Given SUTVA and unconfoundedness, with $W_i$ being the binary treatment indicator for unit $i$, the CATE given representation $\boldsymbol{\phi}_{i, s_1, s_2}$ is defined as:
\begin{align}
\tau\left(\boldsymbol{\phi}_{i, s_1, s_2}\right) &\coloneqq \mathbb{E}\left[Y_i(1) - Y_i(0) \mid \boldsymbol{\phi}_{i, s_1, s_2}\right] \;\; \nonumber \textit{(SUTVA)}\\
&= \mathbb{E}\left[Y_i \mid \boldsymbol{\phi}_{i, s_1, s_2}, W_i=1\right] - \mathbb{E}\left[Y_i \mid \boldsymbol{\phi}_{i, s_1, s_2}, W_i=0\right] \;\;\textit{(Unconfoundedness)}
\end{align}
\noindent Given $\bphi_{i, s_1, s_2}$ and unconfoundedness, CATE can be estimated using a function, $h_{\theta}: \mathbb{R}^{2d'} \to \mathbb{R}$:
$
\hat{\tau}_i = h_{\theta}\left(\boldsymbol{\phi}_{i, s_1, s_2}\right),
$
where $\hat{\tau}_i$ is the estimated CATE for unit $i$ based on the extracted features.

To estimate CATEs, we need to estimate both the representation extraction function $\phi$ and the estimation function $h_{\theta}$. Our goal is to learn a function $f_{\phi}$ that extracts the most causally relevant covariates from the image at multiple levels of representation (e.g., household and village). Having found $f_{\phi}$, we then draw upon the CF approach, a well-established method for estimating $h_{\theta}$ under unconfoundedness \citep{athey2019estimatingtreatmenteffectscausal}. 

With this algorithm for estimating CATEs, we then seek to maximize the heterogeneity signal of different multi-scale representations using a metric, $\mu(\cdot)$, designed to quantify the extent of effect heterogeneity detected given input features. In our case, $\mu(\cdot)$ will denote the RATE Ratio as a principled metric that allows one to evaluate model performance without ground truth individual treatment effects, overcoming the unobservability of true CATE values \citep{holland1986statisticsandci, yadlowsky2021evaluating}. For a set of image conditioning variables, $\bM_i$, the RATE is:
\begin{equation}\label{eq:RATE}
  \textrm{RATE} = \int_{0}^1 \alpha(q) \bigg(
  \underbrace{{\E}[Y_i(1)-Y_i(0) \mid F({\tau}(\bM_i)) \geq 1 - q]}_{\textrm{ATE among top $q$-th percentile under rule ${\tau}$}}
  -
  \underbrace{{\E}[Y_i(1) - Y_i(0)]}_{\textrm{Baseline ATE}}  \bigg)\; \textrm{d} q, 
\end{equation} 
A more detailed exposition is in \S\ref{s:RateDetails}. The RATE Ratio (the measure $\mu$) is defined as $\frac{\text{RATE}}{\text{sd(RATE)}}$. Motivation for using the RATE Ratio lies in the fact that (1) it is interpretable and comparable across contexts as a scale-free quantification of heterogeneity signal and (2) it can be used as an asymptotic $t$-statistic of the existence of treatment heterogeneity in the population given the conditioning data (here, pre-treatment satellite image arrays). The detailed algorithm for RATE Ratio estimation is presented in \S\ref{s:RateDetails}. 

\begin{figure}[htb]
    \centering
    \includegraphics[width=0.40\textwidth]{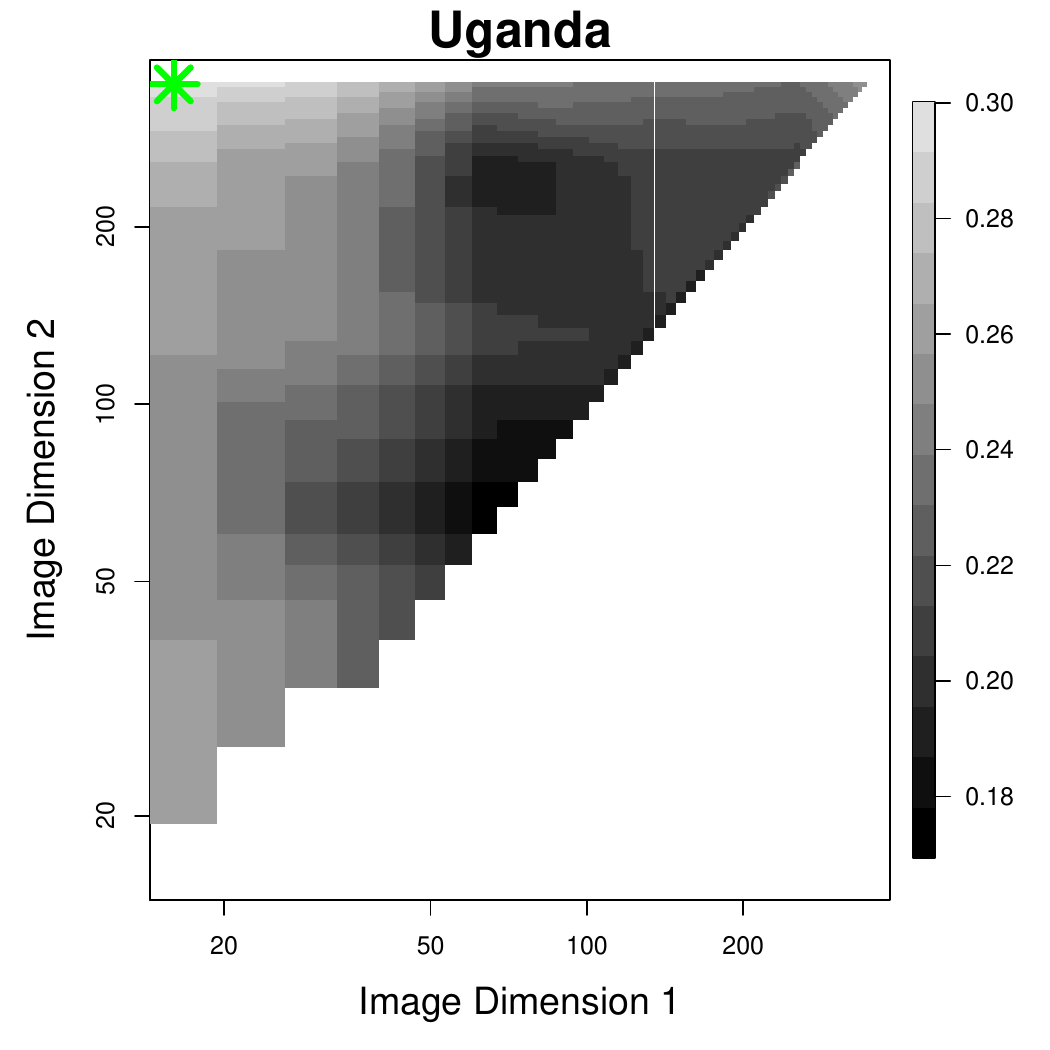}  
    \includegraphics[width=0.40\textwidth]{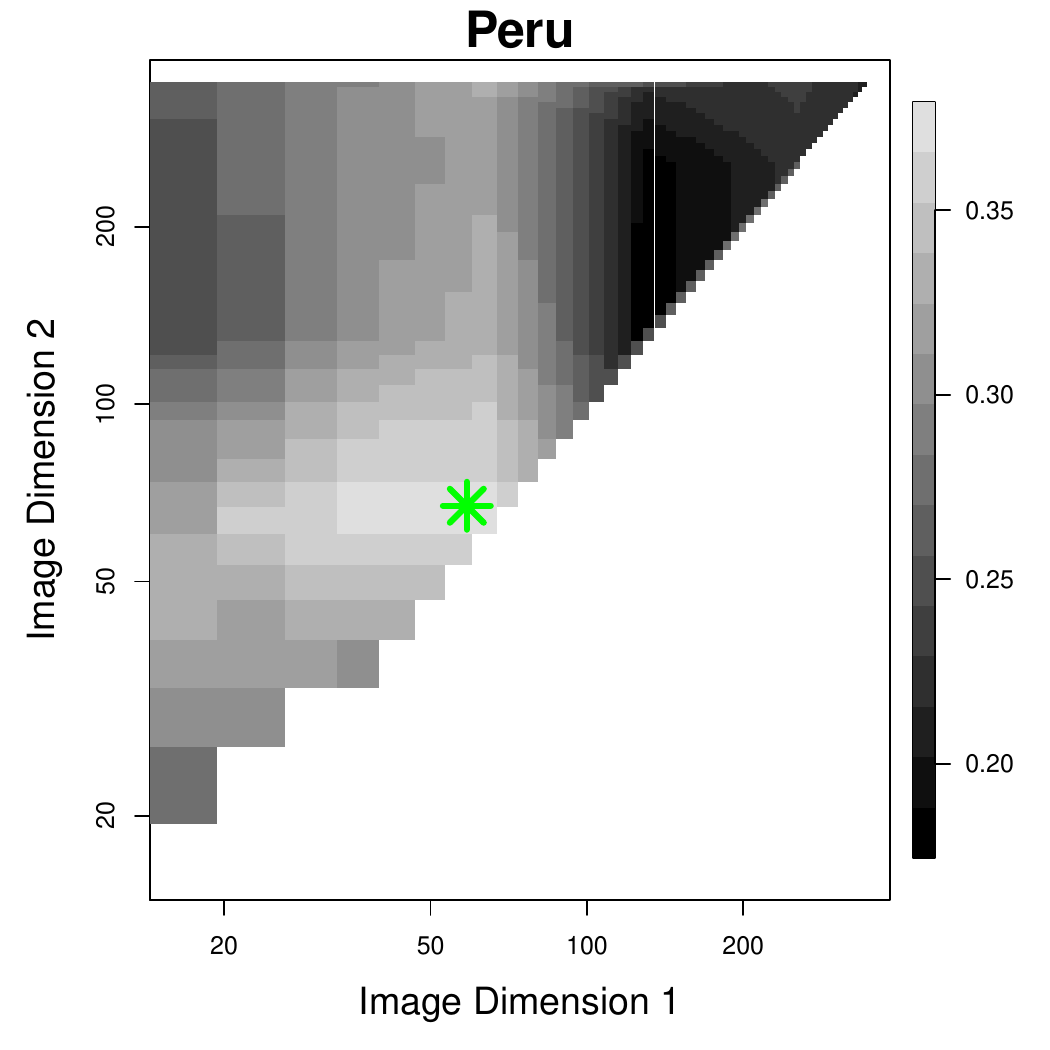}  
    \caption{
Analysis of RATE values for Uganda and Peru RCTs across a range of $s_1$, $s_2$ values (higher/lighter values are better).  
}
    \label{fig:HeatMapMean}
  \end{figure}

\begin{figure}[htb]
    \centering
    \includegraphics[width=0.55\textwidth]{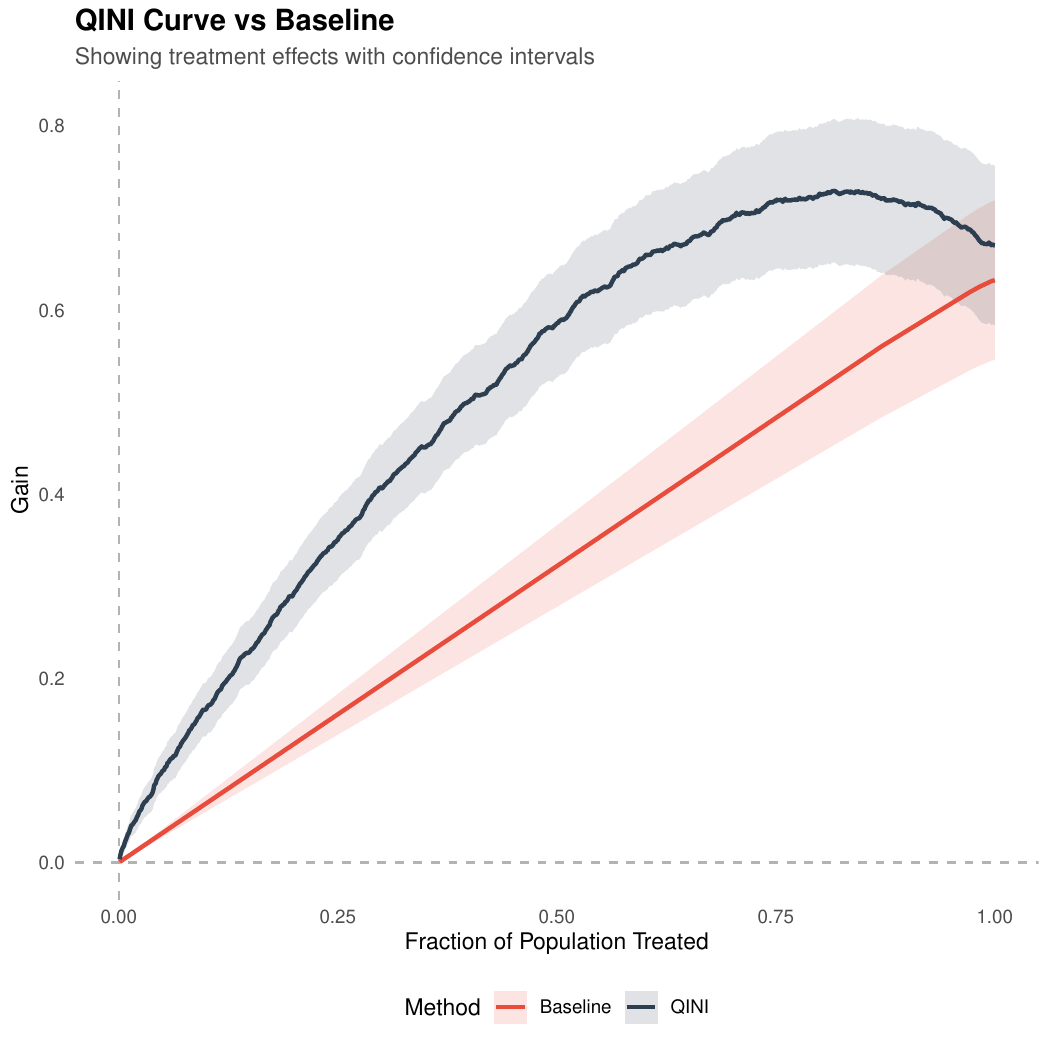}  
    \caption{
Uganda Qini results with CLIP-RSICD representations ($s_1^*=16$, $s_2^*=$349). 
}
\label{fig:QINIPlotsUganda}
  \end{figure}

\begin{figure}[htb]
\centering
\includegraphics[width=0.55\textwidth]{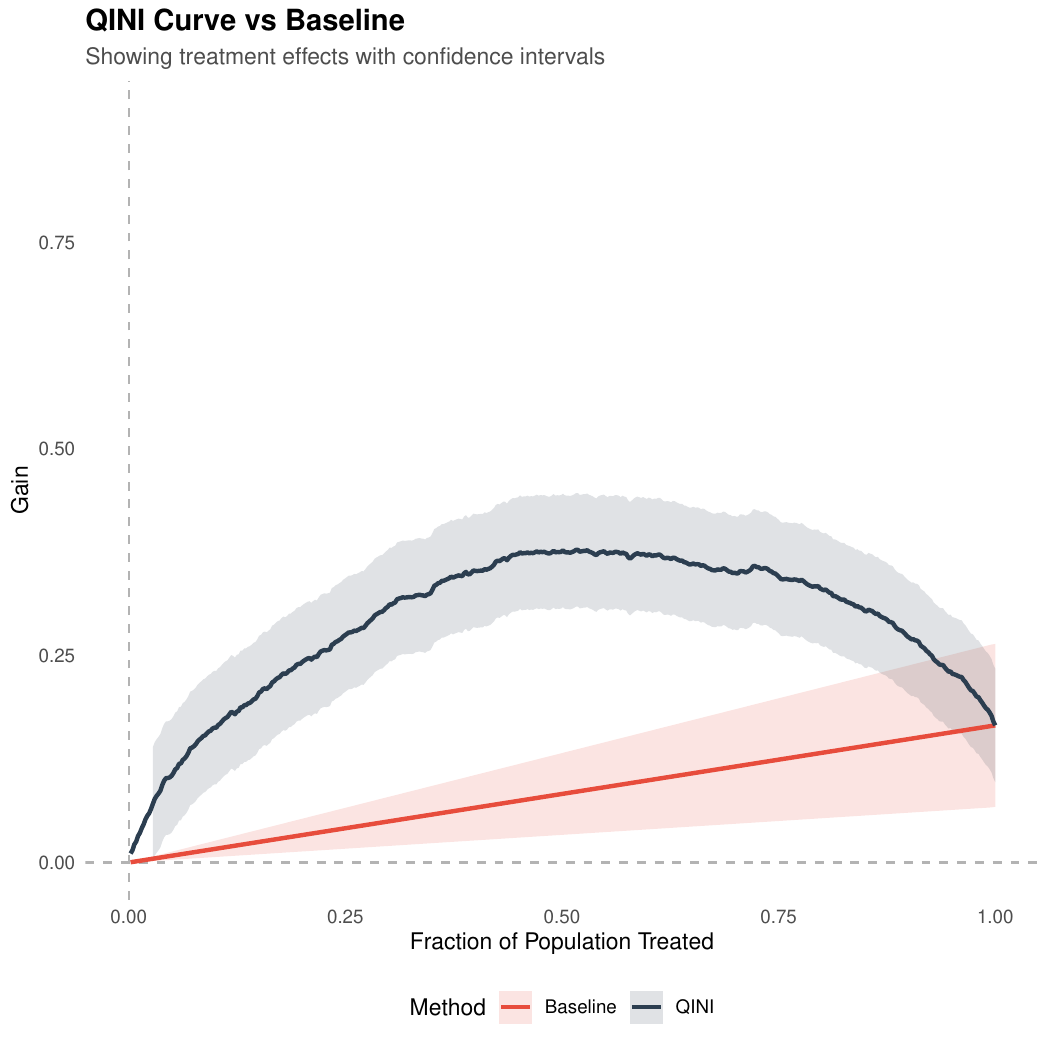}  
    \caption{
Peru Qini results with CLIP-RSICD representations ($s_1^*=$64 and $s_2^*=$64). 
}
\label{fig:QINIPlotsPeru}
  \end{figure}

The RATE Ratio is chosen as the primary metric for model evaluation due to its comparability across RCTs and outcomes. However, RATE Ratio is not as interpretable as the more policy-oriented metrics of RATE and Qini Curves \citep{yadlowsky2021evaluating, sverdrup2024qinicurvesmultiarmedtreatment}. RATE alone is a summary measure of heterogeneity in the data captured by the model and downstream policy relevance of the current heterogeneity model, and Qini Curves plot the gain over randomization of assigning treatment according to CATE estimates from a model. We validate the RATE Ratio metric with the values produced by these two alternatives. We find a high correlation between the three metrics, indicating that in our context RATE Ratio also quantifies the policy benefits of a specific CATE estimation model. Due to the high correlation, we leave RATE and Qini figures for the Appendix (Figure \ref{fig:HeatMapMean}, Figure \ref{fig:QINIPlotsUganda}, Figure \ref{fig:QINIPlotsPeru}). For a discussion of Qini Curves, see \S\ref{s:qinicurve}. 

With this way of generating representations from images, CATEs from representations, and heterogeneity measures from CATEs, we now formalize our optimization when performing multi-scale inference. Our optimization proceeds by comparing a multi-scale heterogeneity signal against a baseline of comparison involving the optimal single-scale-only input. Specifically, we have the following optimization:
\begin{equation}\label{eq:GlobalLoss}
\textrm{ \textit{Goal:}  }  \max_{s_1, s_2} \bigg\{ \E\bigg[ \mu\left( h_{\theta}\left( f_{\phi}\left( \bM_{i, s_1},\, \bM_{i, s_2} \right) \right) \right)\bigg] - \max_{s} \E\bigg[\mu\left( h_{\theta}\left( f_{\phi_s}'\left( \bM_{i, s} \right) \right) \right)\bigg] \bigg\},
\end{equation} 
\noindent Expectations are taken over population variability. The resulting optimized value is $G$, the Multi-scale Gain. While optimization over $s_1$ and $s_2$ does not depend on $s$, we include the term involving $s$ to establish a baseline---if $G$ is negative, we have evidence in favor of using single-scale-only representations.  We use grid search for optimization (see \S\ref{s:Algorithms}).

\section{Simulation}\label{s:Sim}

\begin{figure}[htb]
    \centering
    \begin{minipage}{0.40\textwidth}
        {\scriptsize
        \begin{align*} 
        Y_i \sim 
        \begin{cases} 
        \mathcal{N}(\textrm{mean}=0,\textrm{var} = 0^2) & \text{if $i$ is not perturbed}, \\
        \mathcal{N}(100, 100^2) & \text{if $i$ is {\scshape Masked}}, \\
        \mathcal{N}(-100, 100^2) & \text{if $i$ is {\scshape Edge Faded}}, \\
        \mathcal{N}(100, 100^2) & \text{if $i$ is {\scshape Contrasted}}, \\
        \mathcal{N}(0, 200^2) & \text{if $i$ is {\scshape Masked} and {\scshape Edge Faded}}.
        \end{cases}
        \end{align*}
        }
    \end{minipage}
    \hfill
    \begin{minipage}{0.40\textwidth}
        \centering
        \includegraphics[width=\textwidth]{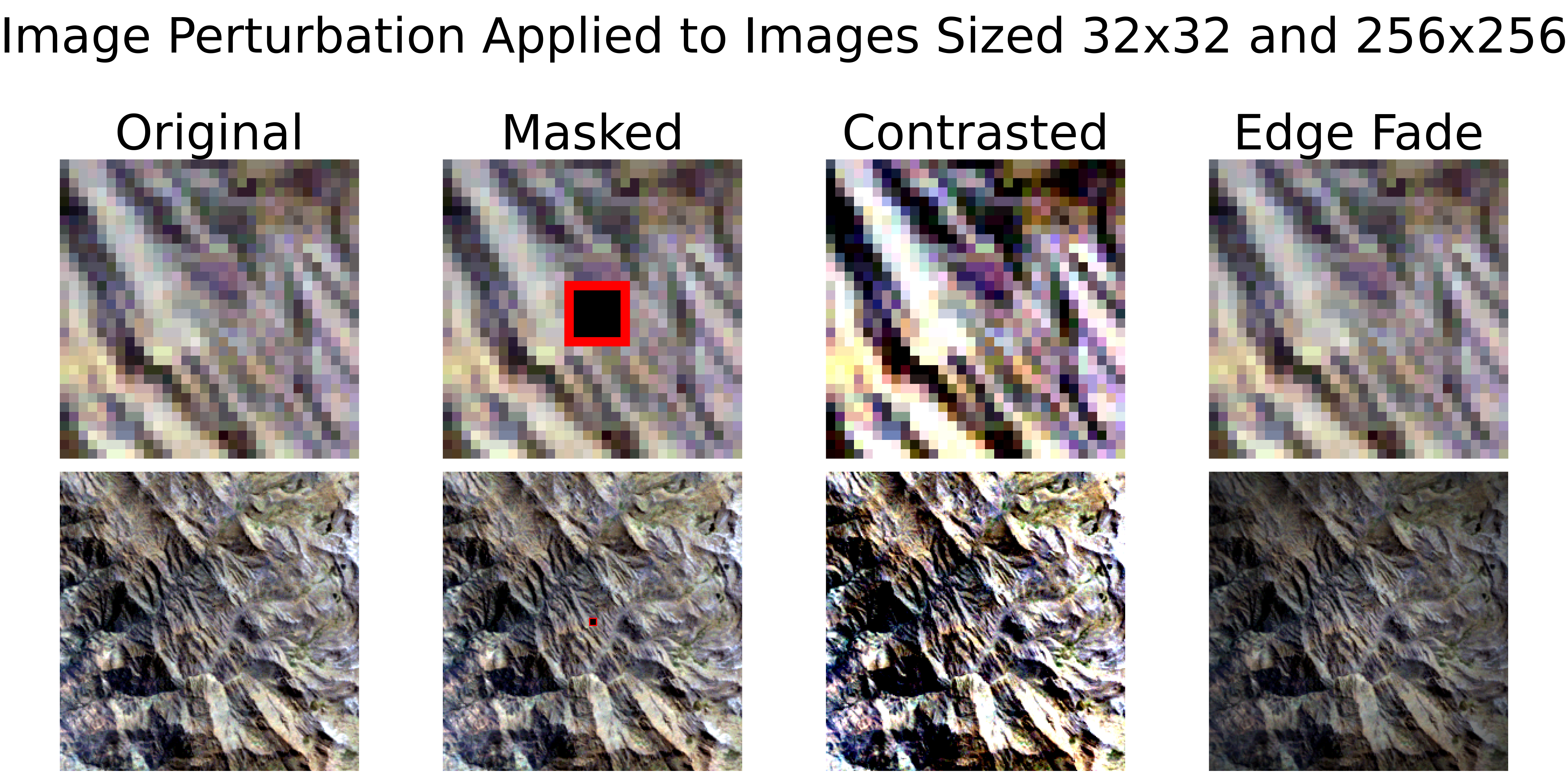}  
    \end{minipage}
    \caption{Visualization of image perturbations ({\scshape Right}) and their corresponding distributions ({\scshape Left}). The equation defines the distribution of $Y_i$ based on the perturbations applied, while the images illustrate these perturbations for different image sizes. Masking used in the simulation experiments is $2\times2$ pixels, enlarged here for visibility.}
    \label{fig:augmentation_vis}
\end{figure}

While the RATE Ratio is a useful theoretical metric, contextual-scale evaluation in an applied context is difficult due to the lack of ground-truth CATE data. In developing our methods, we thus use a simulation to supplement findings from the later RCT analysis. Our simulation employs images of size 32$\times$32 and 256$\times$256 pixels drawn from the Peru RCT. We design three image perturbations corresponding to three scales of causal features (see Figure \ref{fig:augmentation_vis}). We perturb images' \textit{household-level} features by masking the center of the image, \textit{neighborhood-level} features by adding an image fading to the edge of the larger scale image, and \textit{global context features} by applying image contrast. For each of our experiments, we choose a subset of these perturbations, each applied independently to half of the images. The different experiments help us discern how our model will perform when differently scaled covariates are present. Synthetic outcomes are then constructed by adding a deterministic signal with Gaussian noise (see Figure \ref{fig:augmentation_vis}).

In our simulation, we design a setup that mimics an RCT. Each unit has an associated image, and we apply various perturbations---such as masking or edge fading---to these images. The specific type of perturbation applied to a unit's image determines its outcome (and associated CATE). We systematically test the ability of different modeling strategies to detect and leverage these image-based signals in estimation. We measure how well we identify features driving the outcome of interest with the 5-fold cross-validated out-of-sample $R^2$. 

With the outcome and image data defined, we then train a Multi-Layer Perceptron (MLP) on top of representations generated by the CLIP model to predict the outcome of the units. We defer further details to the Appendix (\S\ref{s:SimulationDetails}). 
For a single-scale approach, the input to the MLP is the representation generated by the CLIP image encoder of an image of a fixed size. We then apply Multi-Scale Representation Concatenation with no dimensionality reduction to generate the multi-scale analog.

\begin{figure}[htb]
    \centering
\includegraphics[width=0.5\textwidth]{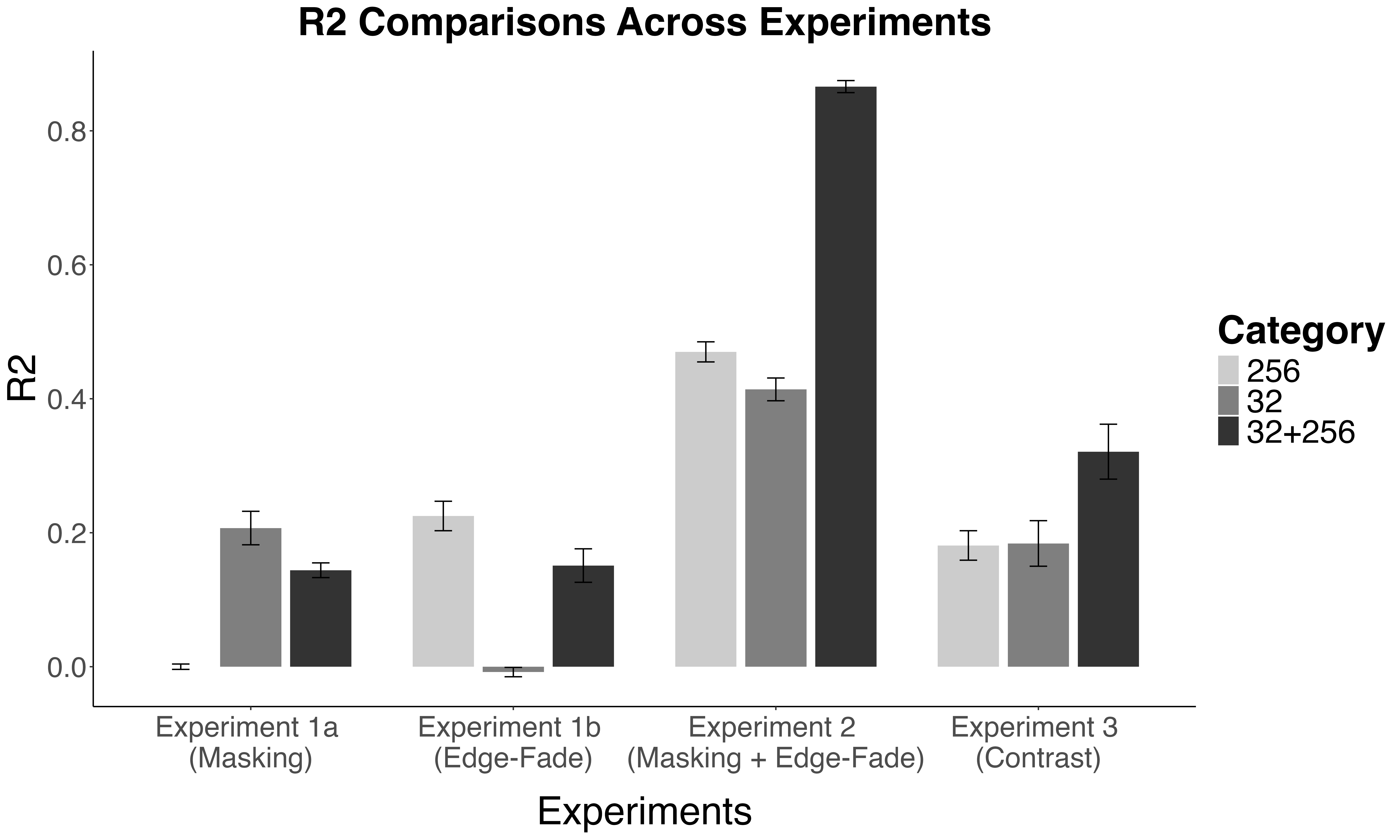}  
    \caption{Experiments 1a and 1b apply household/neighborhood-specific perturbation. Experiment 2 applies household and neighborhood-specific perturbation. Experiment 3 applies global perturbation that has a uniform effect on the whole image. Experiment 1 shows that even if there are no cross-scale effects, Multi-Scale Concatenated procedure can recover much of the signal at one scale. Experiment 2 validates that the procedure can capture multi-level signals if present and outperform a single-scale approach. Experiment 3 suggests that Multi-Scale Representation Concatenation could improve a model's ability to capture predictive signals even when the signal is recoverable from any one scale.}
    \label{fig:Sim_Results}
\end{figure}

Three sets of experiments are performed on the perturbed dataset to explore performance of a multi-scale modeling approach when (a) only household or neighborhood level information is present in the dataset, (b) when both levels of information are present in the dataset, and (c) when the global feature is present in the dataset. We found that our simple concatenation-based multi-scale modeling can recover most of the information present on one level of resolution and captures signal simultaneously from both scales if present. Perhaps more surprisingly, multi-scale modeling also better captures global features (see Figure \ref{fig:Sim_Results}).

While the investigations present here isolate the influence of specific local, intermediate, and global perturbations, we investigate all possible combinations in the Appendix (Table \ref{tab:MultiscaleResultsTable}), where an additional global perturbation of 90-degree image rotation is included. The results accord with those presented in Figure \ref{fig:Sim_Results}.

\section{Application to Anti-Poverty RCTs}

While the simulation results are suggestive of potential benefits of multi-scale analysis, we now turn to quantify its benefits in the context of real RCTs. Our analysis here draws on unique experimental datasets from diverse country contexts: Peru and Uganda. For both, there is evidence that SUTVA is satisfied as spillover effects are determined to be unlikely and treatment implementation is standardized. These datasets therefore provide a solid foundation for exploring the impacts of scale across different societal settings.

\subsection{Data: Treatment, Imagery, and Outcome}

The Peru data are drawn from a program designed to alleviate poverty \citep{banerjee2015}, with treatment occurring from 2007-2014. The outcome is household poverty; treatment is a multi-faceted intervention that combines short-term aid and long-term support. Due to image availability, we use Landsat 5 satellite imagery between 2000 to 2003 with 30 $\times$ 30 $m$ pixels centered around each household, applying a cloud mask and median filter over images, visualized in the Appendix (Figure \ref{fig:SatIm}). 

The Uganda RCT was also designed to reduce poverty, here, by giving young people business grants to improve human capital \citep{blattman2020long}. Treatment occurs from 2008-2012, and data are drawn from the Landsat 7 ETM+ Mosaics from 2000 for their high quality and image resolution (14.5$\times$14.5 $m$ pixels). For Peru, geolocations at the household level are available, whilst for Uganda, geolocations are only available at the village level. 

\begin{figure}[htb]
    \centering
    \includegraphics[width=0.45\textwidth]{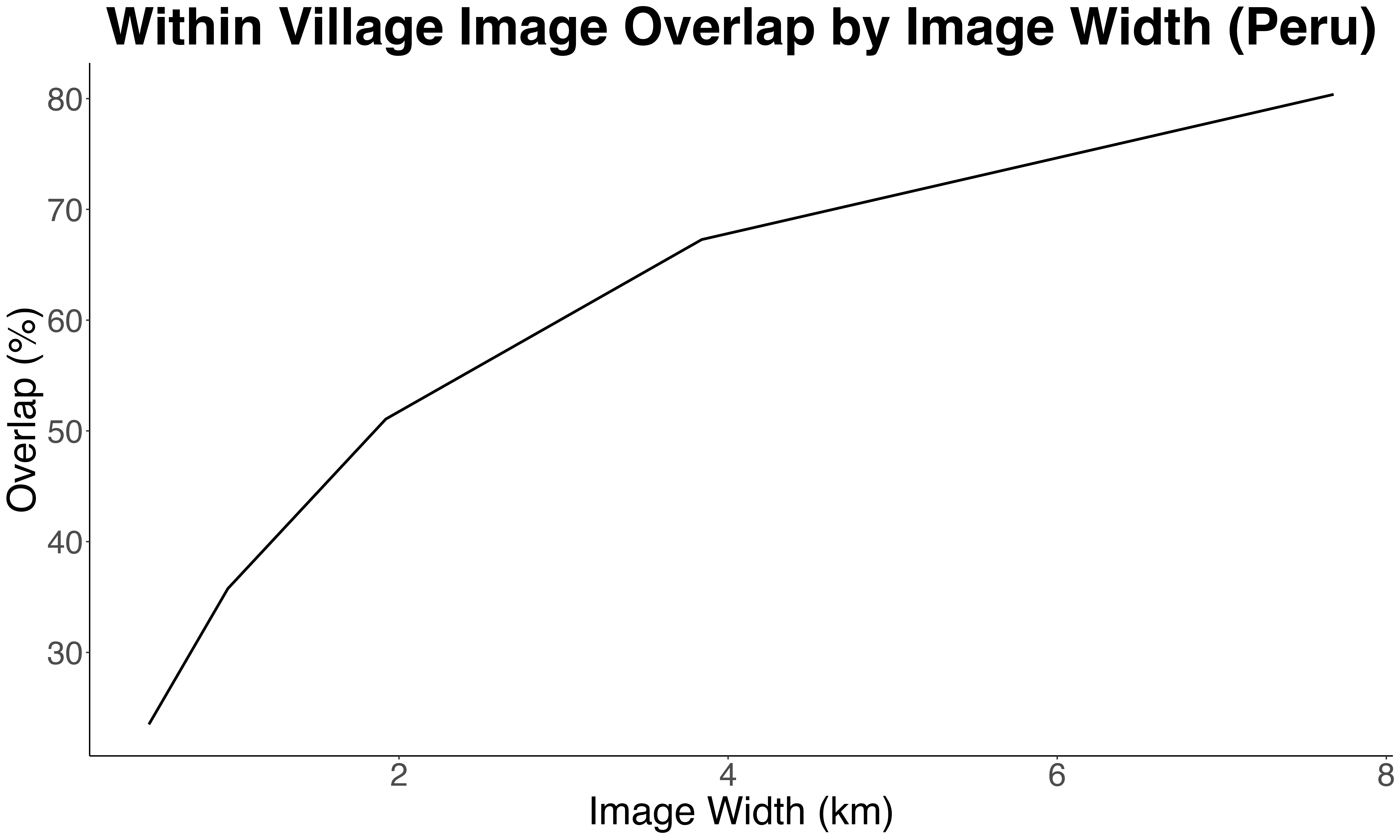}  
    \includegraphics[width=0.45\textwidth]{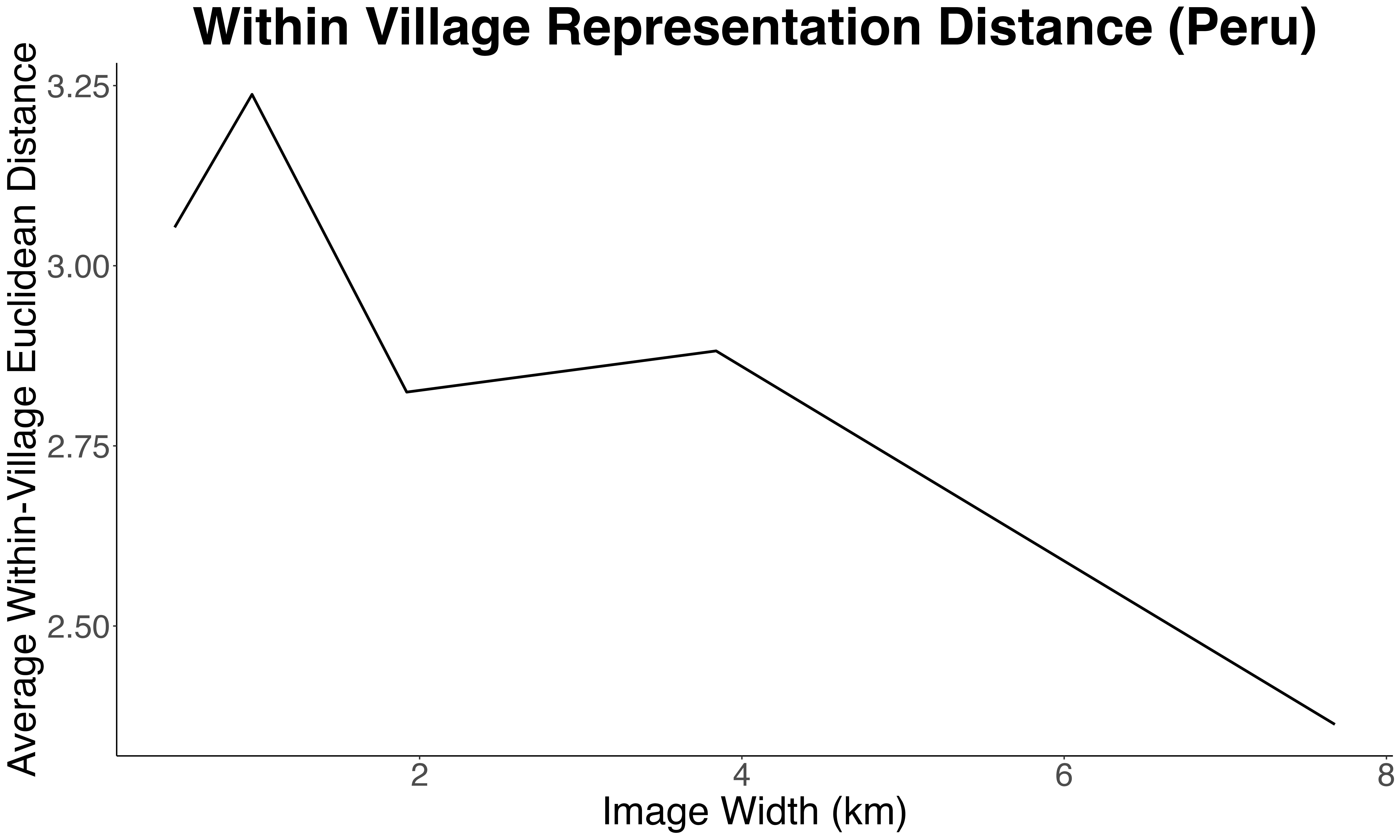} 
    \caption{
     {\scshape Left.} Average pairwise overlap of input images for individuals in a Peru village when image width is varied. {\scshape Right.} For CLIP-RSICD, the mean Euclidean distance of representations inside the same village decreases with increasing image width.
}
    \label{fig:Overlap}
  \end{figure}

Because the Peru geolocations are at the household level, we can analyze the average pairwise image overlap of individuals in the same village. The analysis shows an increase in percentage overlap with image size (left panel of Figure \ref{fig:Overlap}). We further find that image representation distance decreases as the image size increases (right panel), emphasizing the need to use large and small images to obtain heterogeneous representations between individuals inside the same village.

\subsection{Multi-scale RCT Analysis}

We next conduct multi-scale RCT analysis on the two datasets using the CLIP-RSICD image model. Results from the SWIN and Clay model are reported in the Appendix (\S\ref{s:AdditionalEmpirical}). We concatenate CLIP household image representations from images of widths $\{$16, 32, 64, 128, 256, 349$\}$ (i.e., 0.5-10 $km$) and use a CF as CATE estimator. RATE Ratio is used as our evaluation metric, with results for RATE and Qini Curves in the Appendix (\S\ref{s:AdditionalEmpirical}). Due to the composability of our multi-scale procedure, different base image encoders can be chosen for images at different scales. For simplicity, we use the same base encoder throughout. 

Overall, we find that Multi-Scale Representation Concatenation improves or does no worse than single-scale-only analyses. Using a suitable dimensionality reduction function $r$ further increases the performance of a multi-scale procedure. 

\subsubsection{Improvement Over Single-Scale Baseline}

We first analyze the improvement of the multi-scale procedure over the baseline via Equation \ref{eq:GlobalLoss}, where we subtract the optimal single-scale from the optimal multi-scale RATE Ratio to quantify the relative increase in heterogeneity signal. The Multi-scale Gain metric is reported in Table \ref{tab:ClipDiffs}. With benchmark CLIP-RSICD-based analysis, strong evidence of multi-scale heterogeneity is found in Uganda ($G=$
\UgandalcliplrsicdlFALSEMeanDifflpc{} ({\textit s.e.}$=$\UgandalcliplrsicdlFALSEMeanDifflpclse) and
\UgandalcliplrsicdlFALSEMeanDiff{} ({\textit s.e.}$=$\UgandalcliplrsicdlFALSEMeanDifflse)
for PC and raw representations, respectively);
some evidence is found in Peru ($G=$
\PerulcliplrsicdlFALSEMeanDifflpc{} ({\textit s.e.}$=$\PerulcliplrsicdlFALSEMeanDifflpclse) and 
\PerulcliplrsicdlFALSEMeanDiff{} ({\textit s.e.}$=$\PerulcliplrsicdlFALSEMeanDifflse)
for PC and raw representations, respectively).

\begin{table}[!htbp] \centering 
  \caption{RATE ratio differences from Equation \ref{eq:GlobalLoss}.
                                               Standard errors in parentheses.
                               ``clip-rsicd'' an EO fine-tune of CLIP.
                               PC denotes principal component representations.
                               $s^*, s_1^*, s_2^*$ denote optimal image 
                               dimensions in the single- and multi-scale cases using raw (uncompressed) representations. 
                               } 
  \label{tab:ClipDiffs} 
\footnotesize 
\begin{tabular}{@{\extracolsep{5pt}} cccccc} 
\\[-1.8ex]\hline 
\hline \\[-1.8ex] 
PC: Multi-scale Gain & Multi-scale Gain & PC: \{$s^*$\}/\{$s_1^*$, $s_2^*$\} & \{$s^*$\}/\{$s_1^*$, $s_2^*$\} & Case & Model \\ 
\hline \\[-1.8ex] 
0.68 (0.09) & 0.00 (0.08) & \{32\}/\{32, 64\} & \{64\}/\{64, 64\} & Peru & clip-rsicd \\ 
0.95 (0.10) & 0.41 (0.09) & \{349\}/\{32, 128\} & \{16\}/\{16, 349\} & Uganda & clip-rsicd \\ 
\hline \\[-1.8ex] 
\end{tabular} 
\end{table}

Table \ref{tab:ClipDiffs} also provides insight into the optimal image dimensions selected in each data-model-analysis combination. Strikingly, although the largest image context (349 pixels) is the best for single-scale Uganda analysis, when compared with the Multi-scale Concatenated procedures, in no case is the largest image context best---i.e., using the largest image array does not maximize the heterogeneity signal in either of the RCTs. Moreover, smaller images generally generate higher heterogeneity signals than larger images. This hints at some of the challenges in multi-scale analysis, where features from larger images are packed into a fixed-dimension embedding space, leading to the possible degradation of information related to important household- or village-level dynamics. 

\subsubsection{Analysis of RCT Results}

\begin{figure}[htb]
    \centering
    \includegraphics[width=0.35\textwidth]{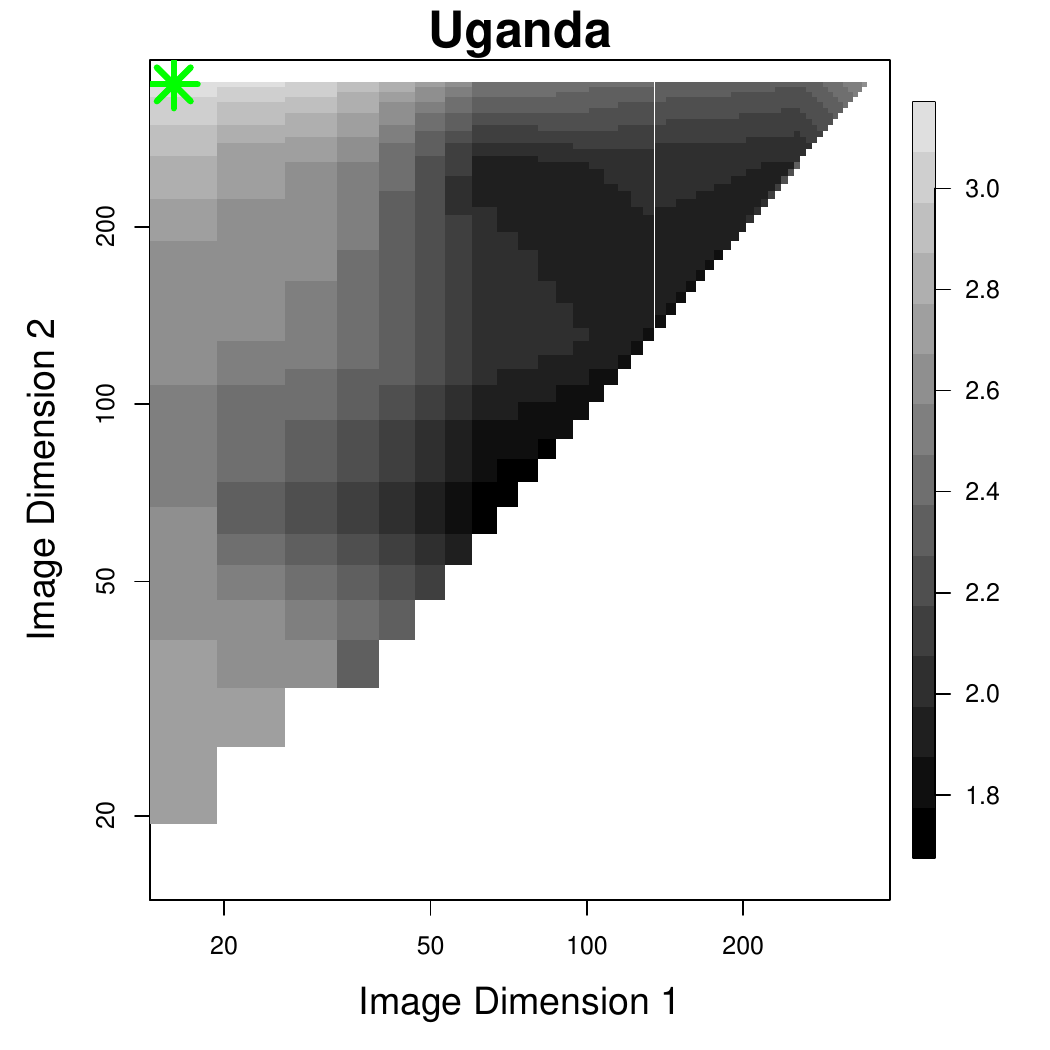}  
    \includegraphics[width=0.35\textwidth]{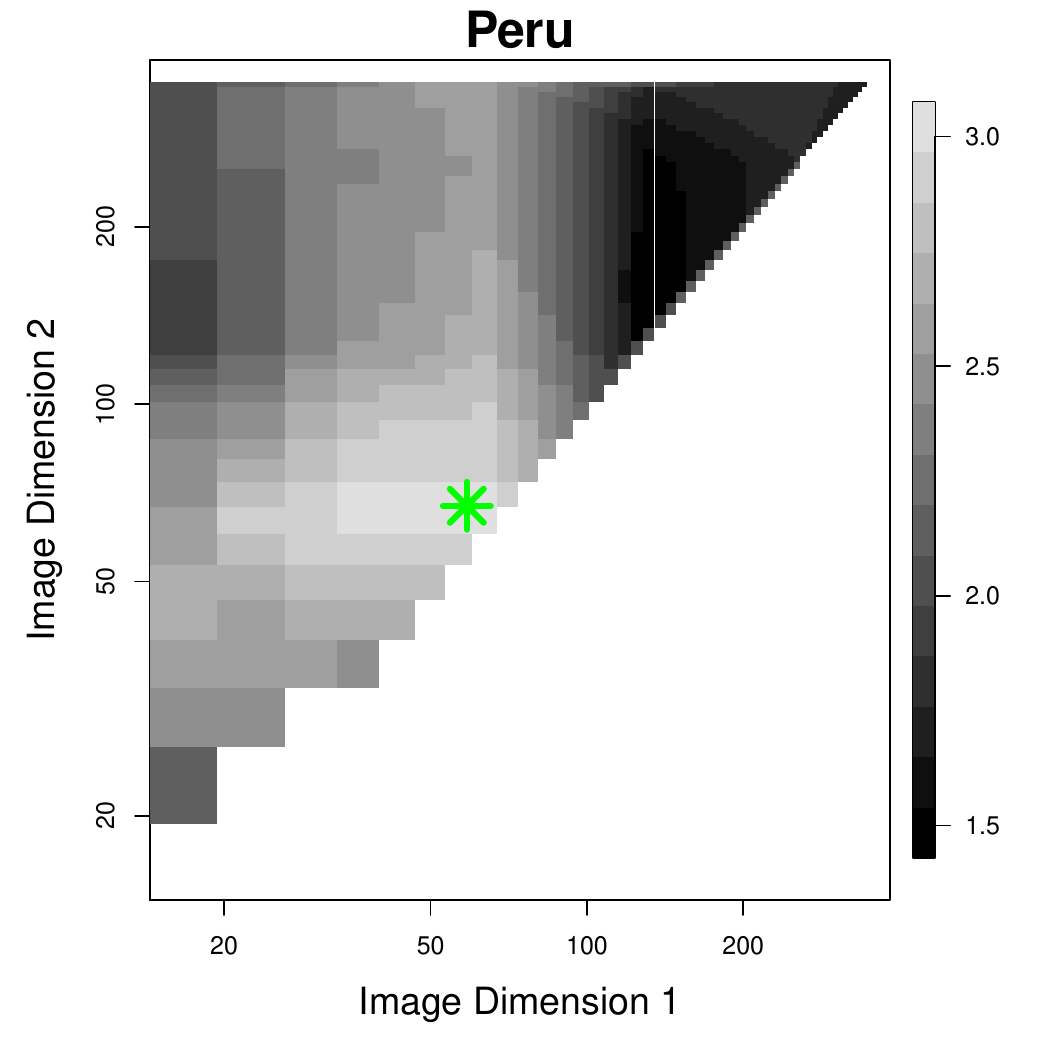}  
    \caption{
Analysis of RATE Ratios for Uganda and Peru RCTs across a range of $s_1$, $s_2$ values (higher/lighter values are better). We see that the maximum heterogeneity signal in both RCTs is detected with small/medium-sized $s_1$ ($\sim$64) and large $s_2$ ($\sim$350). $*$ indicates optimal selection. See Figure \ref{fig:HeatMapMean} for analysis using RATE values (instead of ratios). 
}
    \label{fig:HeatMap}
  \end{figure}

\noindent With further analysis, we find more evidence for the use of multi-scale information by the heterogeneity model. In Figure \ref{fig:HeatMap}, we see that the optimal RATE Ratio occurs when we concatenate representations from a large image with those from a more localized context. This difference is larger in Uganda than in Peru in terms of raw pixel numbers, but may be partially explained by the higher-resolution of the Uganda images.

\begin{figure}[htb]
    \centering
    \includegraphics[width=0.35\textwidth]{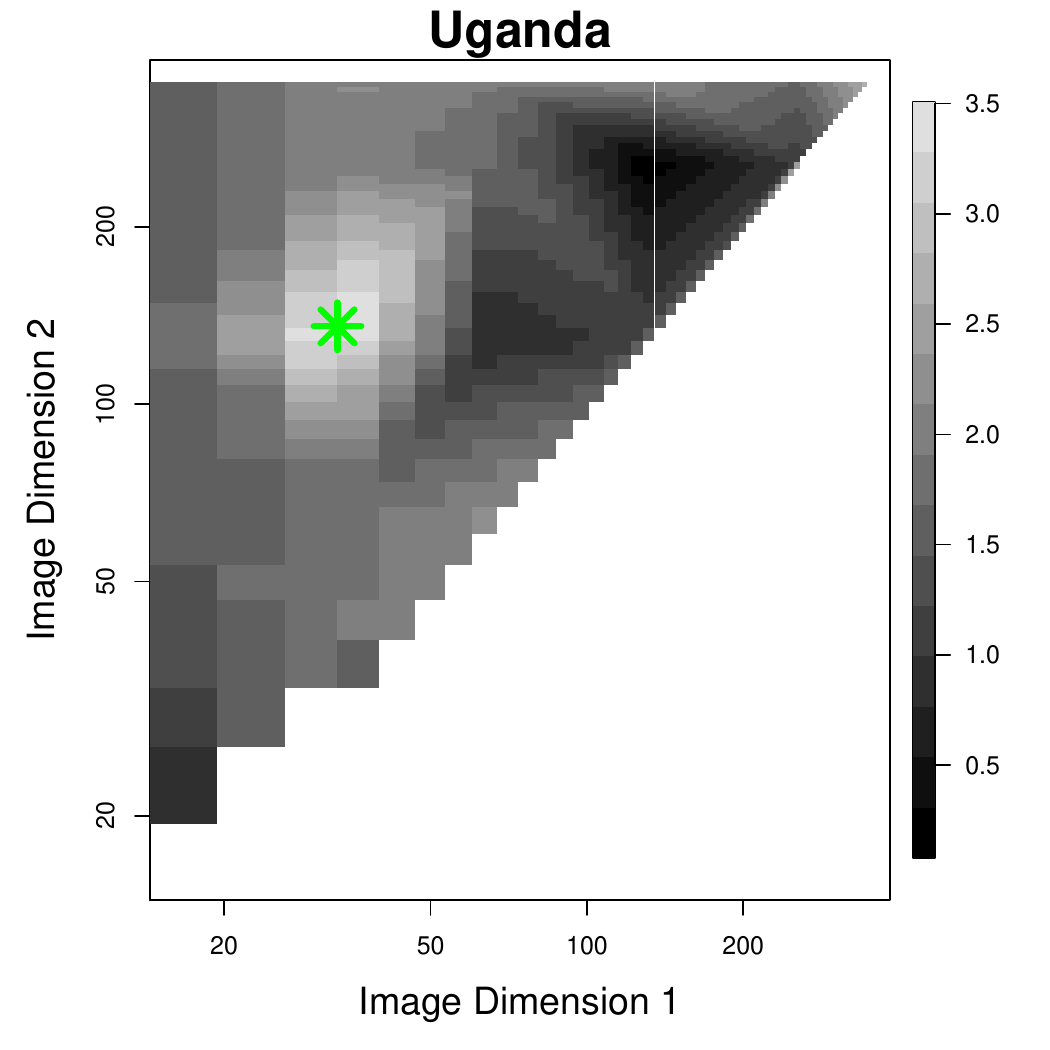}  
    \includegraphics[width=0.35\textwidth]{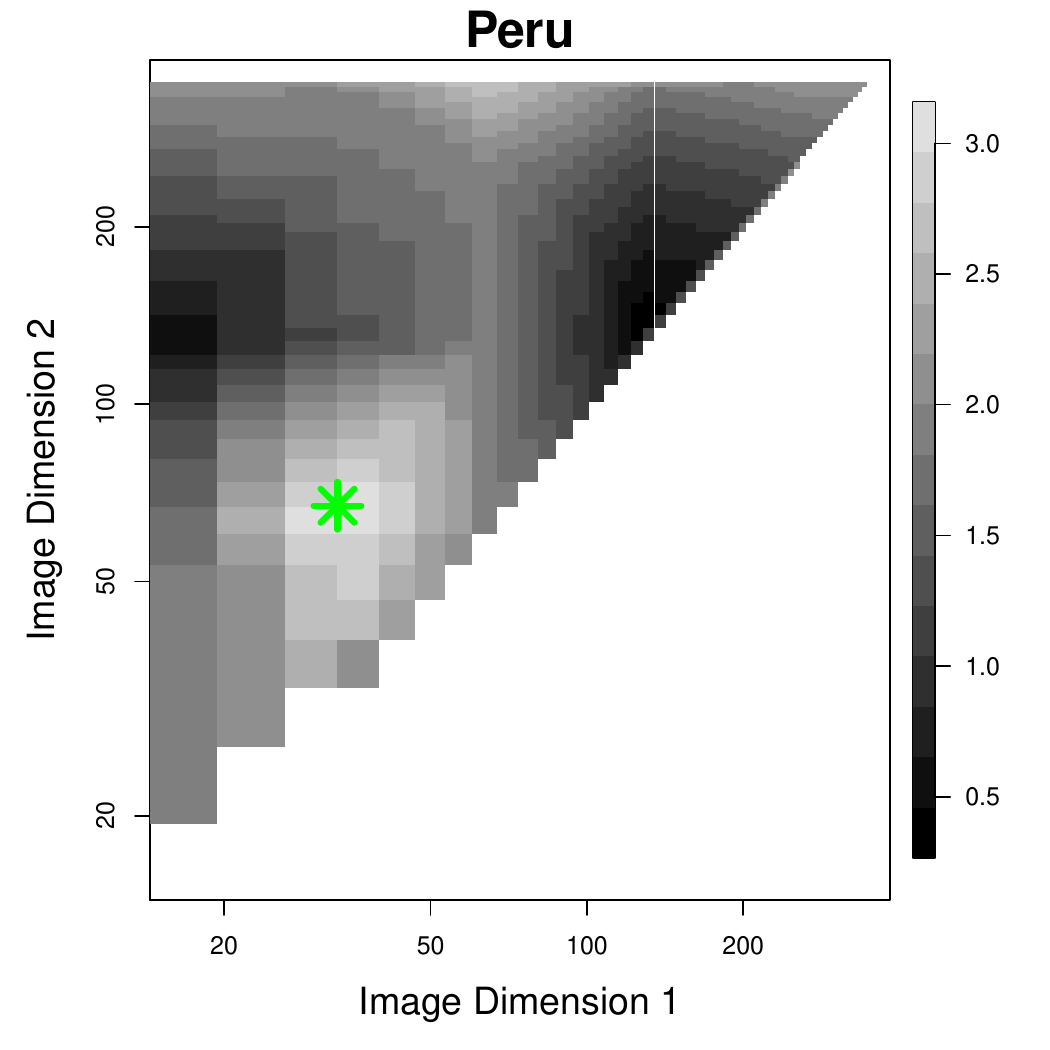}  
    \caption{
Analysis of RATE Ratios across a range of $s_1$, $s_2$ values with PC representations. Higher/lighter values are better. $*$ indicates optimal selection.
}
\label{fig:HeatMapPC}
\end{figure}

Because Multi-Scale Representation Concatenation increases the dimensionality of the input covariates to the CATE estimation model, dimensionality reduction techniques could improve model performance. Figure \ref{fig:HeatMapPC} displays the RATE Ratio heat maps from PC compressions of the neural features (50 dimensions). Using PC representations, we see even stronger evidence for the role of multi-scale representations in improving heterogeneity signal and stronger model performance as evaluated by the RATE Ratios, and the optimality point is shifted towards the center of the heat map. We did not do an exhaustive search for the optimal PC procedure but anticipate that the performance of multi-scale would improve further with further optimization of the dimensionality reduction technique employed.

\begin{figure}[htb]
    \centering
    \includegraphics[width=0.40\textwidth]{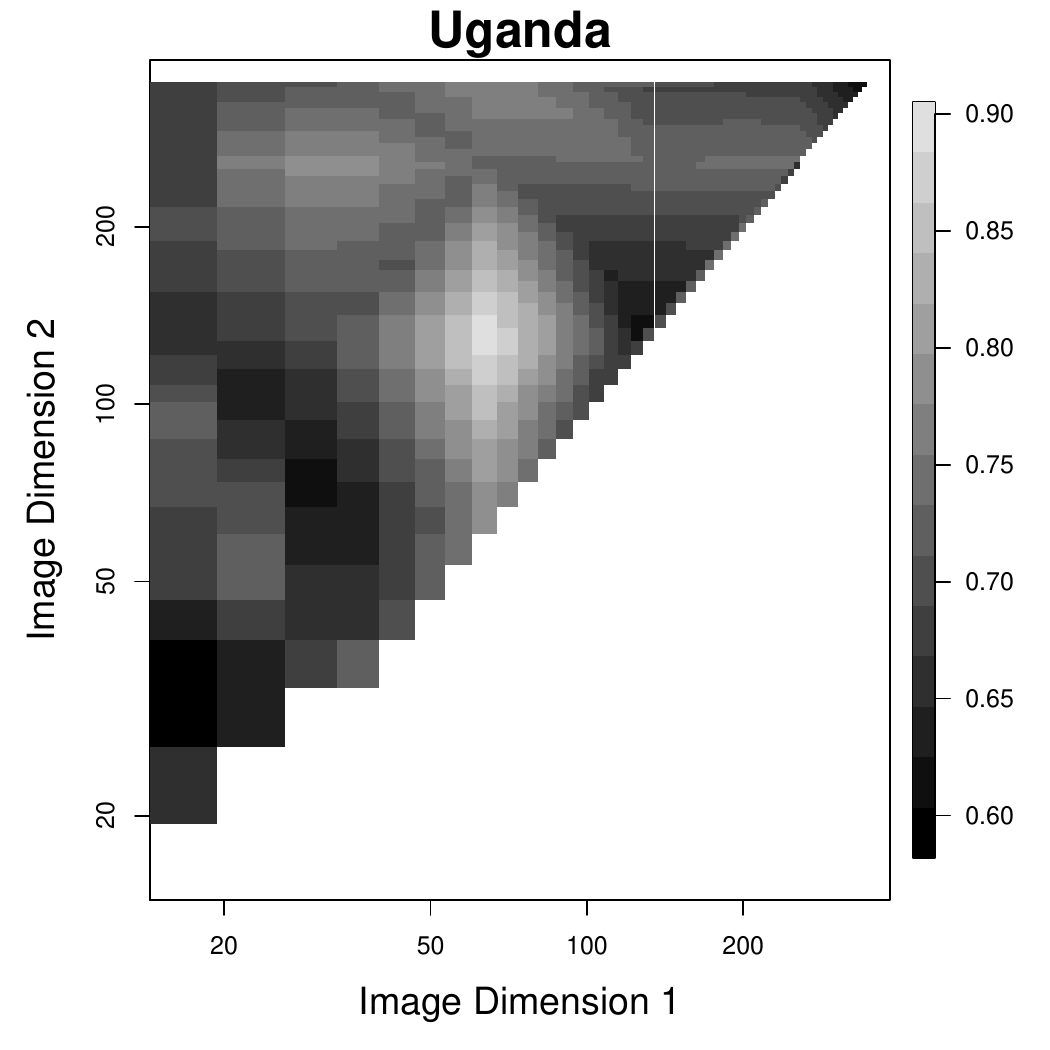}  
    \includegraphics[width=0.40\textwidth]{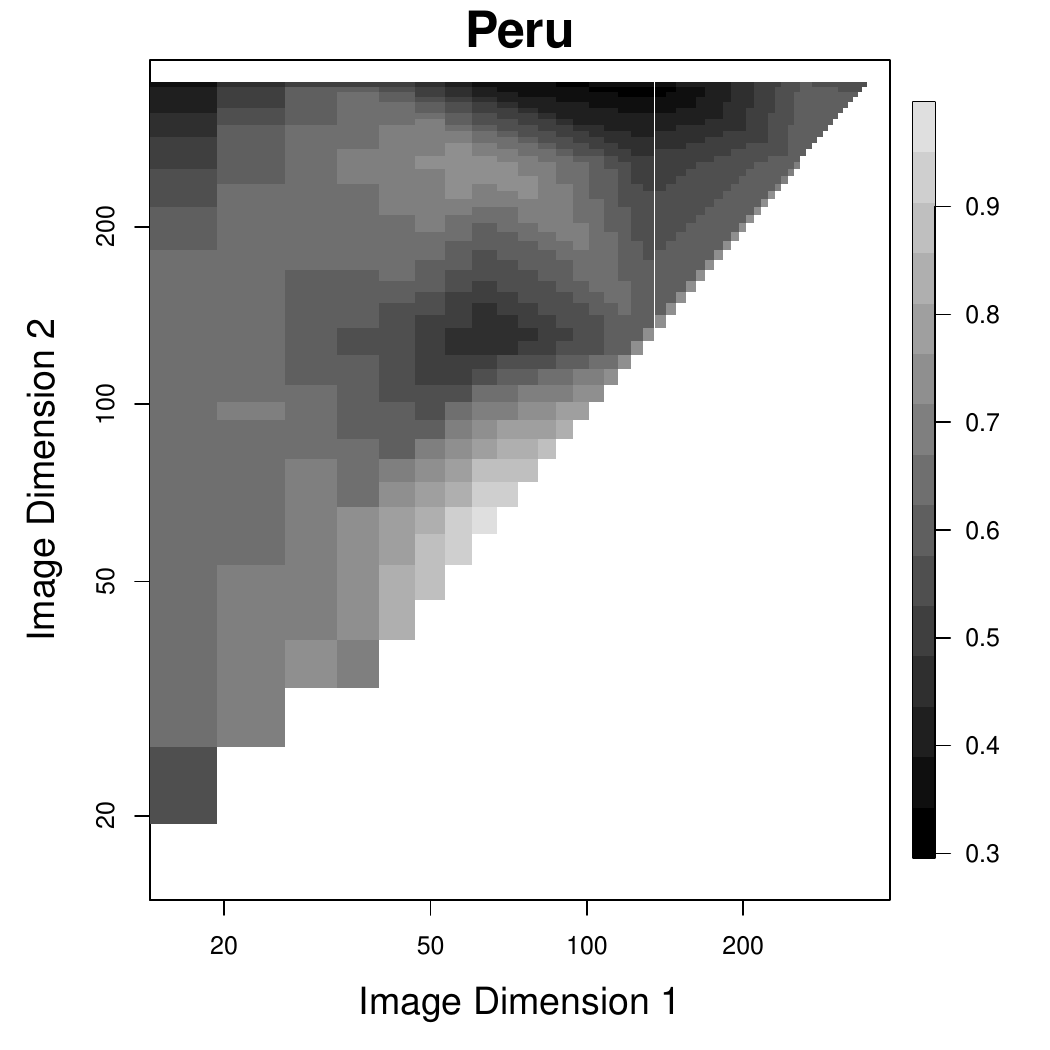}  
    \caption{
Analysis of features predictive of treatment effects. Colors indicate the fraction of the top 10 features (via Causal Forest) predictive of treatment effects from the smaller image scale ($s_1$). We see that, in general, information from both scales is used across the range of $s_1$, $s_2$ values. 
}
\label{fig:HeatInterpret}
\end{figure}

\textbf{Evidence of Multi-scale Use by Heterogeneity Model.}  Due to the interpretability of the concatenation-based multi-scale procedure, we can directly investigate whether there is evidence that the heterogeneity models are incorporating information from multiple scales in forming effect estimates. We probe the utilization of multi-scale information in Figure \ref{fig:HeatInterpret}. Here, we find that the effect heterogeneity model, in general, uses information from both scales in modeling treatment effects. 

We measure the CF model’s information utilization by examining the fraction of the top 10 most informative features (via CF) derived from representations at the smaller scale (i.e., $\boldsymbol{\phi}_{i,s_1}$). Values near 0.5 indicate balanced use of smaller and larger scales; values near 1 indicate dominance of unit-scale features; those near 0 indicate dominance of larger-scale ones. According to this measure, gain from multi-scale information is stronger in Uganda imagery than in Peru imagery, a finding congruent with the analysis of RATE Ratios. A confounding factor here could be the lower resolution of the Peru Imagery (which decreases the amount of information detectable with smaller scales).

\textbf{Multi-scale Dynamics With Weak Prior Information About Unit Locations.} \label{s:WeakPriorInformation} Although we observe an improvement in RATE Ratio in using Multi-Scale Representation Concatenation, the procedure so far described requires strong prior information on the geographic location of households. We therefore investigate performance when we randomly sample image centers so that they are no longer centered around villages/households. To investigate this, we displace household and village locations so that small images no longer contain information from the target units' location and re-run the same analysis as above. We find that there is a decrease in multi-scale dynamics for raw representations (for Uganda, a weak-prior-information $G$ of \UgandalcliplrsicdlTRUEMeanDiff{} ({\textit s.e.}$=$\UgandalcliplrsicdlTRUEMeanDifflse), for Peru, a difference of
\PerulcliplrsicdlTRUEMeanDiff{} ({\textit s.e.}$=$\PerulcliplrsicdlTRUEMeanDifflse)), but not for PC representations (for Uganda, a $G$ of
\UgandalcliplrsicdlTRUEMeanDifflpc{} ({\textit s.e.}$=$\UgandalcliplrsicdlTRUEMeanDifflpclse); for 
Peru, 
\PerulcliplrsicdlTRUEMeanDifflpc{} ({\textit s.e.}$=$\PerulcliplrsicdlTRUEMeanDifflpclse) and 
for PC and raw representations, respectively). 

\begin{figure}[htb]
    \centering
    \includegraphics[width=0.40\textwidth]{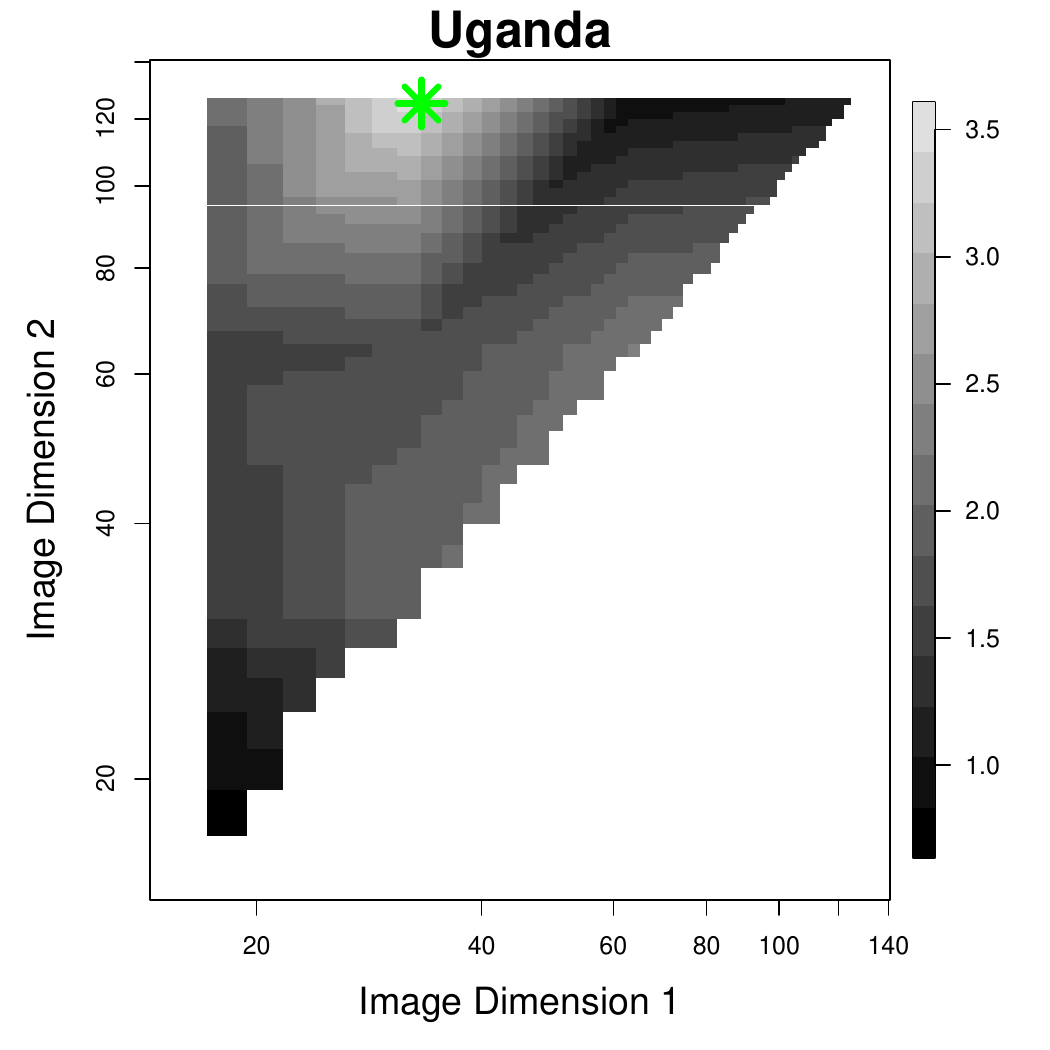}  
    \includegraphics[width=0.40\textwidth]{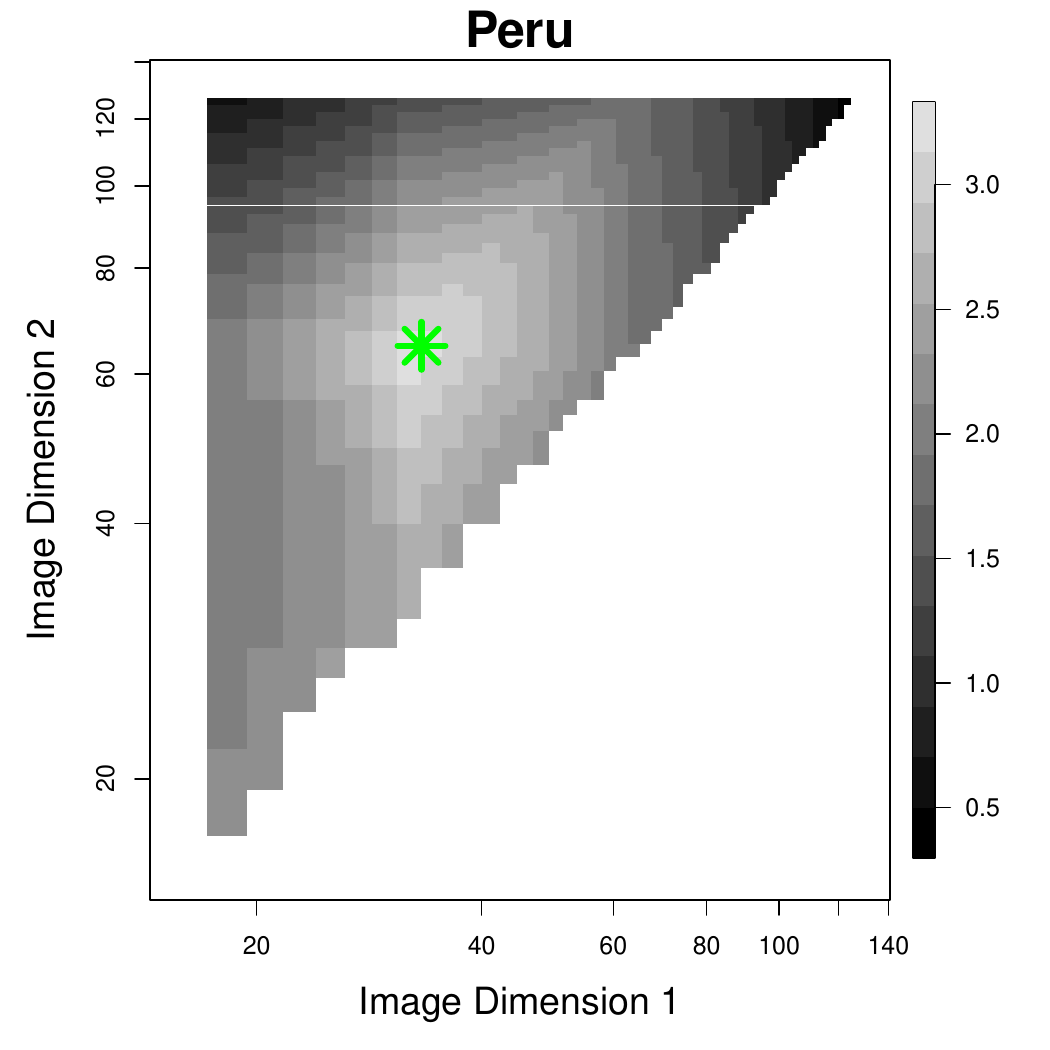}  
    \caption{
Analysis of RATE Ratios for Uganda and Peru with displaced image centers across a range of $s_1$, $s_2$ values (higher/lighter values are better). $*$ indicates optimal selection. 
}
    \label{fig:HeatMapDisplaced}
  \end{figure}

More broadly, Figure \ref{fig:HeatMapDisplaced} shows that, even with only weak information about unit locations, multi-scale analysis can improve signals about heterogeneity; scale information is not specific to household-level features but also other features about localized contexts. We also find that heterogeneity signals are actually higher when using displaced images than the actual images around units (mean RATE Ratio of \AveDisplaced{} vs. \AveNonDisplaced{}), averaging across all method and data choices. In other words, even without household-specific knowledge, heterogeneity signals can still propagate through the integration of multi-scale features into effect heterogeneity modeling. We investigated this in further simulations and found that multi-scale analysis improves signals about heterogeneity, although less than the case with strong prior information (see Table \ref{tab:MultiscaleResultsTable_perturbed} in the Appendix).

\subsection{Scaling Scales}\label{s:ScalingScale}

\begin{figure}[htb]
    \centering
    \includegraphics[width=0.40\textwidth]{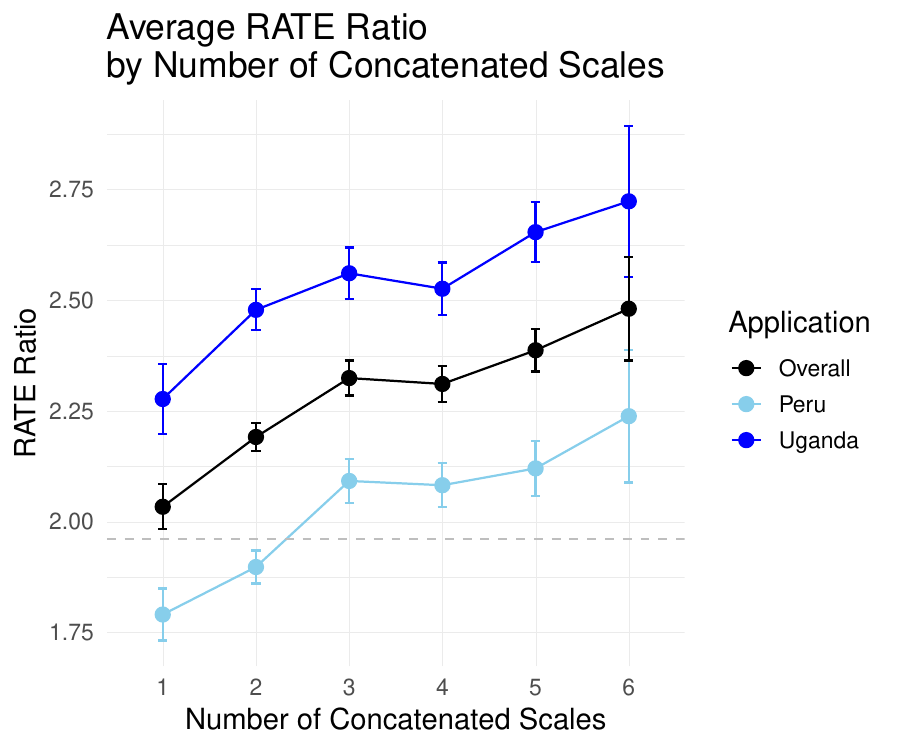}  
    \caption{
As the number of scales grows, so too does average RATE ratio performance. 
The dotted grey line marks 1.96, which represents the 5\% significance threshold for detecting heterogeneity (assuming asymptotic normality).
}
    \label{fig:ScalingScaling}
  \end{figure}

We next consider whether any scaling regularities emerge as the number of concatenated scales grows. Figure \ref{fig:ScalingScaling} investigates dynamics related to the number of scales employed. The average RATE Ratio increases with the number of concatenated scales, indicating that incorporating information from multiple scales enhances the model's ability to detect treatment effect heterogeneity. The RATE Ratio rises from approximately 2 for a single scale to a peak of around 2.5 when five scales are concatenated. Beyond this point, performance stabilizes, as evidenced by the RATE Ratio remaining at 2.5 for six scales. This suggests that concatenating up to five scales captures increasingly diverse information related to household-specific features and broader contextual dynamics to improve the heterogeneity signal, though additional scales beyond this threshold yield diminishing returns due to potential redundancy. This pattern is consistent across Peru and Uganda RCTs. 

  Overall, we see that Multi-Scale Representation Concatenation improves the performance of deep learning models in EO-based CATE estimation, whether information is strong or weak about the exact location of observational units. Moreover, Multi-Scale Representation Concatenation is interpretable in that investigators can readily probe the relative performance of different multi-scale choices on heterogeneity signal---noting that, in comparison, many of the multi-scale inference strategies described in the computer vision literature up to this point are, to a point, black box, and do not offer a ready interpretation of which scales matter, and which do not. Multi-Scale Representation Concatenation thus introduces a composable procedure that allows any single-scale procedure to integrate heterogeneity information from small and large images to better capture both individual heterogeneity and neighborhood-level effects, allowing for more informative causal inferences.

\section{Discussion}

\textbf{Larger Images Are Not Always Better.} A counterintuitive finding from our analyses is that using the largest possible satellite image context (349 pixels) never maximizes the heterogeneity signal in any of our experimental conditions. This challenges the intuitive assumption that incorporating more contextual information necessarily improves causal inference. We find this ``bigger is not better'' principle manifests in three ways.

First, in single-scale analyses with unperturbed geolocations, smaller images often generate elevated heterogeneity signals, likely because they better capture household-specific characteristics without dilution from broader contexts. Second, even in multi-scale analyses, the optimal combinations typically pair a small or medium-sized image (around 64 pixels) with a large but not maximum-sized image, rather than using the largest available scale. Third, multi-scale analysis can provide for robust inferences even when only approximate location information is available (as evidenced by the displaced locations analysis).

These findings have implications for EO-based causal inference. Rather than defaulting to the largest possible image context, researchers should carefully consider the trade-off between capturing relevant local heterogeneity and broader contextual information, either employing Multi-Scale Representation Concatenation or incorporating multi-scale dynamics into the inferential procedure in other ways to be robust to the intricacies of EO data. 

\textbf{Towards Adaptive Multi-Scale Representation Concatenation.} Despite grid search's promising performance, its computational requirement is exponential in $C$, the number of scales considered for combined analysis. Further, the current Multi-Scale Representation Concatenation technique limits itself to a fixed image-scale combination for each observational unit, whilst in practice the optimal scale for each observational unit may differ.

Our finding that Multi-Scale Representation Concatenation outperforms single-scale analysis suggests that there may be a way to optimize Multi-Scale Representation Concatenation further by estimating an additional function $l: \mathcal{M} \to \mathbb{R}^2$ that takes in a large image over an entire region around an experimental unit and outputs locations $\mathbf{x_i}$ where an image at each scale is centered. This would involve evaluating each geographic location for suitability of analysis, where dense prediction techniques could prove helpful \citep{zuo2022denseprediction}. We present the estimation of $l$ as an open problem.

\textbf{Multi-scale dynamics beyond EO.} Multi-scale dynamics are present in diverse geospatial, biological, and public health contexts. Beyond CATE estimation, these dynamics are relevant for environmental cost-benefit analysis \citep{Druckenmiller2024}, EO-based poverty imputation \citep{burke2021using}, and confounder control---where the latent geospatial confounding structure may operate with multi-scale dynamics \citep{jerzak2022estimating}. We invite further work to incorporate multi-scale dynamics in these diverse settings.

\section{Limitations \& Conclusion}

This paper addressed the methodological challenge of capturing multi-scale dynamics in EO-based causal inference. By combining representations across scales, our approach captures both fine-grained individual-level details and broader contextual information, enhancing the estimation of CATEs. Simulation studies and analysis of two RCTs demonstrate the promise of multi-scale inference in outperforming single-scale-only methods when effect heterogeneity information exists at multiple levels. This offers a promising solution to finding a trade-off between individual heterogeneity and neighborhood-level context, and contributes to the growing literature on deep learning for causal inference \citep{daoudStatisticalModelingThree2023}. 

Limitations remain. First, the approach here assumes SUTVA and unconfoundedness for identification, and thus, an extension to observational inference requires additional assumptions. Second, images used in our study have low resolution, and resolution likely interacts with the gain from doing multi-scale analysis. Third, using high-resolution images in a multi-scale approach raises privacy concerns. Fourth, the approach taken to assess the validity of Multi-Scale Representation Concatenation is empirical; theoretical results would strengthen our understanding of the method. Fifth, the approach is currently only validated on anti-poverty RCT datasets. Further research could be done to address these limitations of the general Multi-Scale Representation Concatenation approach. \hfill $\square$ 

\newpage 
\acks{We thank members of the AI \& Global Development Lab for helpful feedback, resources, and inspiration. We thank Dean Karlan and Andre Nickow for assistance in accessing data for the Peru experiment. We thank Hannah Druckenmiller, Antonio Linero, and SayedMorteza Malaekeh for helpful discussions. All remaining limitations are our own. 
}

\bibliography{mybib}

\section{Appendix} 

\subsection{Data Availability Statement}

Simulation code is made available on GitHub:
\begin{itemize}
\item[] \href{https://github.com/AIandGlobalDevelopmentLab/MultiScaler}{\url{GitHub.com/AIandGlobalDevelopmentLab/MultiScaler}}
\end{itemize}
Uganda replication data are available at: 
\begin{itemize}
\item[] \href{https://doi.org/10.7910/DVN/O8XOSF}{\url{doi.org/10.7910/DVN/O8XOSF}}
\end{itemize}
For privacy reasons, given household-level geolocations in Peru, we cannot make that RCT data available at this time. 

\subsection{Satellite Imagery Processing Details}\label{s:satelliteimageryprocessing}

\begin{figure}[htb]
    \centering
\includegraphics[width=0.4\textwidth]{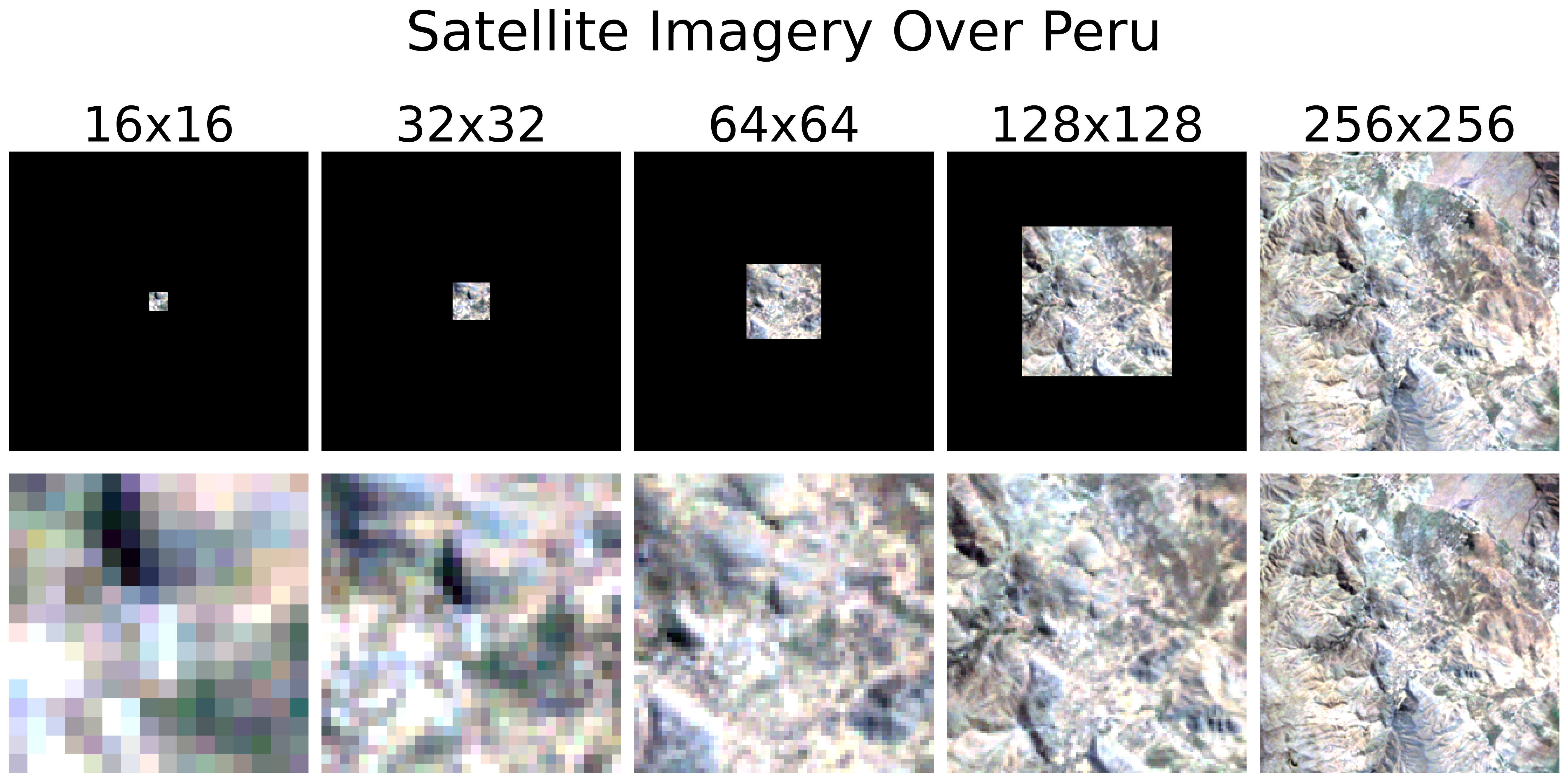}
    \caption{Peru images size 16$\times$16 pixels to 256$\times$256 pixels from $2007$. Image resolution is held at $30$m the highest resolution for the Landsat $5$ and $7$ satellites. Visualized imagery are not centered around RCT participants for privacy considerations. The top row shows images without resizing across changing dimensions; bottom panel shows resized images. Images of different scales are obtained in identical fashion in Uganda.}
    \label{fig:SatIm}
\end{figure}

Direct image cropping is used to obtain satellite imagery at different scales. The effect of this cropping is shown in Figure \ref{fig:SatIm}.

\subsection{RATE Ratio Details}\label{s:RateDetails}

The RATE Ratio is calculated through sample splitting, with $\hat{\tau}$ estimated from half of the samples, and $\hat{\E}$ from the other half \citep{yadlowsky2021evaluating}. The value of $\text{sd(RATE)}$ is estimated through half-sample bootstrap. In Equation \ref{eq:RATE}, $\hat{\tau}(\bM_i)$ is used as a prioritization rule, with $\hat{\E}[Y_i(1)-Y_i(0) \mid F(\hat{\tau}(\bM_i)) \geq 1 - q]$ being the Average Treatment Effect (ATE) among the top $q$-th percentile of treatment respondents as estimated by ${\bM_i}$. $\E[Y_i(1) - Y_i(0)]$ is the baseline ATE, and the difference $\E[Y_i(1)-Y_i(0) \mid F(\hat{\tau}(\bM_i)) \geq 1 - q] - \E[Y_i(1) - Y_i(0)]$ represents the gain in ATE in the respondents in the top $q$-th percentile of estimated CATE over the general population. Finally, this difference is weighed through $\alpha(q)$ and integrated to produce a scalar output.

There are at least two weighting functions under which the RATE Ratio ($\frac{\text{RATE mean}}{\text{RATE SD}}$) has hypothesis testing guarantees in detecting heterogeneity in a population---i.e., $\alpha_{AUTOC}(q) = 1$ and $\alpha_{QINI}(q) = q$. In our context, we report the AUTOC weighting function, which gives more weight to individuals with high response in the integration. This weighting is more relevant for policy analysis when not all individuals will be treated. Indeed, in our study, where the focus is on anti-poverty interventions like cash transfers, identifying households with the largest treatment effects is often a priority for efficient resource allocation, hence the choice of AUTOC weighting over the equi-weighting approach taken by QINI (where all respondents contribute uniformly to the RATE calculation).

\subsection{Related Work} \label{s:relatedwork}

There is a tradition of research into multi-scale phenomena in Computer Vision that has informed our approach to developing methodologies for EO-based Causal Inference. We survey them below.

\textbf{Multi-scale Strategies in Deep Vision Networks.} The success of deep vision networks has been attributed to their ability to learn hierarchical levels of abstractions present in variously-scaled images \citep{szegedy2014going, lecun2015deep}. Deep residual networks enabled by skip connections are multi-scale feature extractors \citep{Jiao2023MultiscaleRepresentationLearning}, and interpretable architectures that take advantage of part-whole hierarchies have been proposed \citep{hinton2021representpartwholehierarchiesneural}. The SWIN Transformer and MLP-mixer are recent architectures that place the hierarchical multi-scale nature of images at the center of their architecture, achieving efficiency and interoperability, respectively  \citep{liu2021swintransformerhierarchicalvision, tolstikhin2021mlpmixerallmlparchitecturevision}. 

\textbf{Explicit Multi-scale Representation Learning in Computer Vision.} A drawback to the above multi-scale approach is that they do not explicitly preserve representations at multiple scales, whilst in EO-contexts such information may be needed (e.g. exact details of the rooftop of a household's home that take up a small number of pixels but encode important information about their socio-economic status). Thus, more relevant to our application scenario are strategies that explicitly preserve lower-scale information. Region Proposal Networks achieve efficient object detection by processing an image globally before identifying local sub-regions \citep{girshick2014richfeaturehierarchiesaccurate, girshick2015fastrcnn, ren2016fasterrcnnrealtimeobject}. Multi-scale residual networks combine residual connections with convolution kernels of varying sizes to adaptively detect features at different scales \citep{Li_2018_ECCV}. In EO-based Deep Learning, the Scale-MAE architecture tackles the scale dependence of geospatial representations by combining Laplacian Pyramids and Ground Sample Distance Encoding with a Transformers-based Masked Auto-encoder framework \citep{reed2023scalemaescaleawaremaskedautoencoder}. Our work adds to this existing literature by introducing a composable and interpretable procedure that adapts arbitrary EO-based causal effect estimation algorithms to incorporate multi-scale dynamics in effect heterogeneity estimation, enabling practitioners to improve their pre-existing effect heterogeneity estimation pipeline with limited additional data requirement and model design effort.

\textbf{Scale Invariant Vision Network.} Scale-invariance has been recognized as an important characteristic of the natural world and a desirable property for vision networks \citep{chater1999scale, han_scale_2020}. Image augmentation through cropping and resizing are commonly used to encourage scale-invariance in models \citep{He2015, yang2023imagedataaugmentationdeep}, and architectures with inductive biases for scale-invariance have been proposed
\citep{chen2016attentionscalescaleawaresemantic, gong2014multiscaleorderlesspoolingdeep, jansson_scale-invariant_2022}. In the EO setting, where input scale and resolution differ markedly depending on modeling choices, designing architectures adaptable to different scales grows in importance \citep{reed2023scalemaescaleawaremaskedautoencoder}. Our work employs Multi-Scale Representation Concatenation to extract features amenable to detection at different scales, providing a general procedure to enable extraction of arbitrarily scaled features in satellite imagery.

\subsection{Qini Curve Details}\label{s:qinicurve}

Our exposition deals with Qini Curves with a single binary treatment and constant costs for each unit. There are generalizations of Qini Curves with varying costs and multiple treatment arms \citep{sverdrup2024qinicurvesmultiarmedtreatment}, but they are not applicable to our RCT data. 

Qini Curves employ the identical sample splitting strategy to obtain the estimated CATEs as with RATE. With the obtained CATEs, Qini Curves construct a policy function $\pi_B(M_i)$, which assigns treatment to a unit $M_i$ if $\hat{\tau}(\bM_i)$ greater than $0$ and in the top $B\%$ of the CATE predictions. It then plots a curve with $B$ as the X-axis and the expected gain in the outcome of interest when one targets the most responsive units according to $\pi_B(M_i)$ as the Y-axis. It is often plotted with a baseline ATE curve which shows the expected gain when the treatment is random assignment.

Both Qini Curves and RATE measure the degree of treatment effect heterogeneity that a model is able to capture from the input covariates. Qini curves are useful under the setting of cost-benefit analysis as its y-axis directly measures the expected gain. 

We implement Qini Curves through the \texttt{maq} package. Doubly robust scores obtained through Causal Forests using the \texttt{grf} package are used as CATE estimators, the cost parameter is set as the constant 1, and 200 bootstrap replications are used for standard error calculation, 

\subsection{Simulation Details}\label{s:SimulationDetails}

Fold and perturbation indices are generated from simple random samples from available indices. We found that making outcomes Gaussian (rather than deterministic) had no statistically significant effect on model performance. We also found that the representations learned by the model are not always robust to small perturbations. By applying contrast image perturbation on top of other image perturbations to experiment 1 without changing the outcome, $R^2$ decreased by $55\%$. This provides motivation for developing model architectures or fine-tuning procedures that are more robust to noise in multi-scale features.

We used a Multi-Layer Perceptron (MLP) with three linear layers, each followed by ReLU activation. The input layer connects to a hidden layer of 128 neurons, followed by a second bottleneck layer of 32 neurons and an output layer providing a scalar prediction. The architecture is mathematically described as follows:
\begin{align*} 
f(\mathbf{x}) = \sigma(W_3 \cdot \sigma(W_2 \cdot \sigma(W_1 \mathbf{x} + b_1) + b_2) + b_3)
\end{align*} 
where \(W_i\) and \(b_i\) represent layer weights and biases, and \(\sigma\) is the ReLU activation.

For the Multi-scale Concatenated model, input dimensionality is doubled, as two image representations are concatenated. We experimented with Principal Component Analysis (PCA) for dimensionality reduction; using only subsets of principal components often led to large performance degradations. 

The {\scshape Edge Fade} image perturbation is implemented using a radial distance-based mask:
\[
\texttt{mask} = \texttt{Clip}(1 - \texttt{distance} \times \texttt{fade\_size}, 0, 1),
\]
and the {\scshape Contrast} image perturbation is applied by transforming each pixel value using,
\[
M_{i,w,h,b}^{{\text{New}}} = \overline{M}_b + c \times (M_{i,w,h,b} - \overline{M}_b),
\]
where \(M_{i,w,h,b}\) is the original pixel value, \(\overline{M}_{b}\) is the mean band intensity, and $c$ scales the contrast.

Here is the full results for the simulations when taking all possible combinations of image perturbations. We additionally tested Rotation as a global context feature.

\begin{table}[!htbp] \centering 
\begin{footnotesize}
  \caption{Simulation $R^2$ for All Combinations of Perturbation Parameters, Values are presented as mean (lower CI, upper CI). R is rotation, M is mask, C is contrast, E is edge.} 
\label{tab:MultiscaleResultsTable} 
\begin{tabular}{@{\extracolsep{5pt}} D{.}{.}{-3} D{.}{.}{-3} D{.}{.}{-3} D{.}{.}{-3} D{.}{.}{-3} D{.}{.}{-3} D{.}{.}{-3} } 
    \hline\hline \\[-1.8ex]  
    R & M & C & E & $256+32$ & $256$  & $32$ \\ 
    \hline \\[-1.2ex]        
0 & 0 & 0 & 1 & \multicolumn{1}{c}{0.152 (0.139, 0.165)} & \multicolumn{1}{c}{0.198 (0.183, 0.212)} & \multicolumn{1}{c}{-0.007 (-0.011, -0.003)} \\ 
0 & 0 & 1 & 0 & \multicolumn{1}{c}{0.295 (0.270, 0.320)} & \multicolumn{1}{c}{0.169 (0.153, 0.184)} & \multicolumn{1}{c}{0.198 (0.193, 0.202)} \\ 
0 & 0 & 1 & 1 & \multicolumn{1}{c}{0.134 (0.124, 0.144)} & \multicolumn{1}{c}{0.097 (0.094, 0.099)} & \multicolumn{1}{c}{0.064 (0.058, 0.069)} \\ 
0 & 1 & 0 & 0 & \multicolumn{1}{c}{0.146 (0.133, 0.160)} & \multicolumn{1}{c}{0.001 (-0.001, 0.003)} & \multicolumn{1}{c}{0.201 (0.191, 0.210)} \\ 
0 & 1 & 0 & 1 & \multicolumn{1}{c}{0.136 (0.122, 0.150)} & \multicolumn{1}{c}{0.086 (0.078, 0.094)} & \multicolumn{1}{c}{0.078 (0.066, 0.091)} \\ 
0 & 1 & 1 & 0 & \multicolumn{1}{c}{0.103 (0.068, 0.137)} & \multicolumn{1}{c}{0.048 (0.041, 0.056)} & \multicolumn{1}{c}{0.088 (0.082, 0.094)} \\ 
0 & 1 & 1 & 1 & \multicolumn{1}{c}{0.082 (0.074, 0.090)} & \multicolumn{1}{c}{0.035 (0.027, 0.044)} & \multicolumn{1}{c}{0.038 (0.031, 0.045)} \\ 
1 & 0 & 0 & 0 & \multicolumn{1}{c}{0.014 (0.009, 0.019)} & \multicolumn{1}{c}{0.010 (0.008, 0.013)} & \multicolumn{1}{c}{0.001 (-0.000, 0.003)} \\ 
1 & 0 & 0 & 1 & \multicolumn{1}{c}{0.646 (0.636, 0.656)} & \multicolumn{1}{c}{0.565 (0.555, 0.576)} & \multicolumn{1}{c}{0.115 (0.098, 0.131)} \\ 
1 & 0 & 1 & 0 & \multicolumn{1}{c}{0.558 (0.550, 0.566)} & \multicolumn{1}{c}{0.511 (0.498, 0.524)} & \multicolumn{1}{c}{0.440 (0.424, 0.455)} \\ 
1 & 0 & 1 & 1 & \multicolumn{1}{c}{0.142 (0.125, 0.159)} & \multicolumn{1}{c}{0.131 (0.118, 0.144)} & \multicolumn{1}{c}{0.071 (0.062, 0.081)} \\ 
1 & 1 & 0 & 0 & \multicolumn{1}{c}{0.593 (0.588, 0.598)} & \multicolumn{1}{c}{0.264 (0.248, 0.280)} & \multicolumn{1}{c}{0.430 (0.422, 0.437)} \\ 
1 & 1 & 0 & 1 & \multicolumn{1}{c}{0.110 (0.102, 0.119)} & \multicolumn{1}{c}{0.074 (0.065, 0.083)} & \multicolumn{1}{c}{0.070 (0.064, 0.076)} \\ 
1 & 1 & 1 & 0 & \multicolumn{1}{c}{0.137 (0.127, 0.148)} & \multicolumn{1}{c}{0.064 (0.059, 0.068)} & \multicolumn{1}{c}{0.097 (0.081, 0.113)} \\ 
1 & 1 & 1 & 1 & \multicolumn{1}{c}{0.072 (0.065, 0.078)} & \multicolumn{1}{c}{0.049 (0.043, 0.055)} & \multicolumn{1}{c}{0.047 (0.041, 0.053)} \\ 
\hline \\[-1.8ex] 
\end{tabular} 
\end{footnotesize}
\end{table} 

The results for the case where there is weak prior information about unit locations is also presented. Under weak prior information, we combine a 256x256 image representation with a 32x32 image drawn uniformly from the image.

\begin{table}[!htbp] \centering 
\begin{footnotesize}
  \caption{Simulation $R^2$ for All Combinations of Perturbation Parameters, Values are presented as mean (lower CI, upper CI). R is rotation, M is mask, C is contrast, E is edge. Weak Prior column combines a 256x256 image representation with a 32x32 image drawn uniformly from the image} 
  \label{tab:MultiscaleResultsTable_perturbed} 
\begin{tabular}{@{\extracolsep{5pt}} D{.}{.}{-3} D{.}{.}{-3} D{.}{.}{-3} D{.}{.}{-3} D{.}{.}{-3} D{.}{.}{-3} D{.}{.}{-3} } 
    \hline\hline \\[-1.8ex]  
    R & M & C & E & $Weak Prior$ & $256$  & $32$ \\ 
    \hline \\[-1.2ex]        
0 & 0 & 0 & 1 & \multicolumn{1}{c}{0.243 (0.197, 0.289)} & \multicolumn{1}{c}{0.195 (0.193, 0.197)} & \multicolumn{1}{c}{-0.006 (-0.008, -0.004)} \\ 
0 & 0 & 1 & 0 & \multicolumn{1}{c}{0.218 (0.188, 0.249)} & \multicolumn{1}{c}{0.193 (0.159, 0.228)} & \multicolumn{1}{c}{0.190 (0.178, 0.203)} \\ 
0 & 0 & 1 & 1 & \multicolumn{1}{c}{0.119 (0.101, 0.136)} & \multicolumn{1}{c}{0.102 (0.094, 0.111)} & \multicolumn{1}{c}{0.065 (0.064, 0.066)} \\ 
0 & 1 & 0 & 0 & \multicolumn{1}{c}{0.002 (0.001, 0.002)} & \multicolumn{1}{c}{0.001 (0.001, 0.001)} & \multicolumn{1}{c}{0.222 (0.175, 0.268)} \\ 
0 & 1 & 0 & 1 & \multicolumn{1}{c}{0.084 (0.072, 0.096)} & \multicolumn{1}{c}{0.073 (0.071, 0.074)} & \multicolumn{1}{c}{0.086 (0.071, 0.100)} \\ 
0 & 1 & 1 & 0 & \multicolumn{1}{c}{0.066 (0.056, 0.076)} & \multicolumn{1}{c}{0.058 (0.052, 0.064)} & \multicolumn{1}{c}{0.093 (0.078, 0.108)} \\ 
0 & 1 & 1 & 1 & \multicolumn{1}{c}{0.053 (0.042, 0.065)} & \multicolumn{1}{c}{0.041 (0.027, 0.056)} & \multicolumn{1}{c}{0.024 (-0.006, 0.054)} \\ 
1 & 0 & 0 & 0 & \multicolumn{1}{c}{0.013 (0.010, 0.016)} & \multicolumn{1}{c}{0.008 (0.005, 0.011)} & \multicolumn{1}{c}{0.000 (-0.001, 0.002)} \\ 
1 & 0 & 0 & 1 & \multicolumn{1}{c}{0.578 (0.559, 0.596)} & \multicolumn{1}{c}{0.550 (0.525, 0.575)} & \multicolumn{1}{c}{0.100 (0.097, 0.103)} \\ 
1 & 0 & 1 & 0 & \multicolumn{1}{c}{0.544 (0.522, 0.566)} & \multicolumn{1}{c}{0.522 (0.501, 0.543)} & \multicolumn{1}{c}{0.428 (0.415, 0.441)} \\ 
1 & 0 & 1 & 1 & \multicolumn{1}{c}{0.126 (0.099, 0.154)} & \multicolumn{1}{c}{0.122 (0.106, 0.138)} & \multicolumn{1}{c}{0.073 (0.059, 0.087)} \\ 
1 & 1 & 0 & 0 & \multicolumn{1}{c}{0.249 (0.225, 0.273)} & \multicolumn{1}{c}{0.276 (0.268, 0.283)} & \multicolumn{1}{c}{0.425 (0.417, 0.432)} \\ 
1 & 1 & 0 & 1 & \multicolumn{1}{c}{0.079 (0.063, 0.094)} & \multicolumn{1}{c}{0.070 (0.064, 0.076)} & \multicolumn{1}{c}{0.071 (0.064, 0.078)} \\ 
1 & 1 & 1 & 0 & \multicolumn{1}{c}{0.065 (0.054, 0.076)} & \multicolumn{1}{c}{0.062 (0.028, 0.096)} & \multicolumn{1}{c}{0.107 (0.074, 0.141)} \\ 
1 & 1 & 1 & 1 & \multicolumn{1}{c}{0.054 (0.046, 0.061)} & \multicolumn{1}{c}{0.051 (0.046, 0.056)} & \multicolumn{1}{c}{0.046 (0.036, 0.055)} \\ 
\hline \\[-1.8ex] 
\end{tabular} 
\end{footnotesize}
\end{table} 

The simulation replication code is in the $\text{Sim}$ folder of the code repository.

\newpage 
\subsection{Algorithms}\label{s:Algorithms}

\begin{algorithm}[H]
\SetAlgoLined
\KwIn{
\begin{itemize}
    \item \(i \in \{1, \cdots, n\}\) denotes the index for observational units.
    \item \(\{\bx_i\}_{i=1}^{n}\) the set of locations of those units.
    \item Sets of image sizes \(\mathcal{S}=\{s_1, s_2, \dots, s_{\textrm{Max}}\}\).
    \item Image fetcher, \(f_I(\bx_i, s)\), that obtains an image centered at a given location \(\bx_i\) with size \(s\).
    \item Image encoders, \(f_{\phi_{s_1}}'\) and \(f_{\phi_{s_2}}'\), that extracts representations from \(\bM_{i,s_1}\), \(\bM_{i,s_2}\).
    \item Dimensionality reduction function $r$ that reduces the dimension of the concatenated representations.
    \item Trainable CATE estimation function \(h_{\theta}(\cdot)\) parametrized by \(\theta\).
    \item Observed outcome of interest \(\mathbf{Y} = (Y_1, Y_2, \dots, Y_n)\).
    \item Binary treatment indicator \(\mathbf{W} = (W_1, W_2, \dots, W_n), W_i \in \{0, 1\}\).
\end{itemize}
}

\KwOut{Optimal image sizes \(s_1^*, s_2^*\), with optimal RATE Ratio*}

\BlankLine
\textbf{Grid Search over \(s_1\) and \(s_2\):}

Initialize \(s_1^* \leftarrow 0\), \(s_2^* \leftarrow 0\)\;

\ForEach{\(s_1 \in \mathcal{S}\)}{
    \ForEach{\(s_2 \in \mathcal{S}\)}{
        Set \texttt{MaxRATERatio} to \(-\infty\)\;

        \ForEach{\(i, \bx_i \in \texttt{enumerate}(\{\bx_i\}_{i=1}^n)\)}{
            \(\bM_{i,s_1} = f_I(\bx_i, s_1)\);\;
            
            \(\bM_{i,s_2} = f_I(\bx_i, s_2)\)\;
        
            \(\boldsymbol{\phi}_{i,s_{1},s_2} = r((f_{\phi_{s_1}'}(\bM_{i,s_1}), f_{\phi_{s_2}'}(\bM_{i,s_2})))\)\;
        }

        Compute \(\widehat{\textrm{RATE Ratio}} = R(\mathbf{W}, \mathbf{Y}, \{\boldsymbol{\phi}_{i,s_{1},s_2}\}_{i=1}^n)\)\;

        \If{\(\widehat{\textrm{RATE Ratio}} > \texttt{MaxRATERatio}\)}{
            \(\texttt{MaxRATERatio} \leftarrow \widehat{\textrm{RATE Ratio}}\)\;
            
            \(s_1^* \leftarrow s_1\)\;
            
            \(s_2^* \leftarrow s_2\)\;
        }
    }
}

\Return{Optimal sizes \(s_1^*\), \(s_2^*\), \texttt{MaxRATERatio}}
\caption{Grid search optimizing multi-scale representations in CATE estimation.}
\end{algorithm}

\subsection{Causal Forest Implementation Detail}\label{s:CausalForest}

We use the \texttt{grf} package in R to implement our Causal Forest. The minimum node size is $5$. The nuisance models are random forests. The number of trees grown is $2000$. 

\subsection{Additional Empirical Results}\label{s:AdditionalEmpirical}

\begin{table}[!htbp] \centering 
  \caption{RATE ratio differences from Equation \ref{eq:GlobalLoss}. 
                               Standard errors in parentheses.
                               ``clay'' denotes the Clay EO foundation model; 
                               ``clip-rsicd'' an EO fine-tune of CLIP; 
                               ``swin'' the SWIN Transformer.
                               PC denotes principal component representations.
                               $s^*, s_1^*, s_2^*$ denote optimal image 
                               dimensions in the single- and multi-scale cases using raw (uncompressed) representations. 
                               } 
  \label{tab:AllDiffs} 
\footnotesize 
\resizebox{\textwidth}{!}{
\begin{tabular}{@{\extracolsep{5pt}} ccccccc} 
\\[-1.8ex]\hline 
\hline \\[-1.8ex] 
PC: Multi-scale Gain & Multi-scale Gain & PC: \{$s^*$\}/\{$s_1^*$, $s_2^*$\} & \{$s^*$\}/\{$s_1^*$, $s_2^*$\} & Case & Model & Displaced? \\ 
\hline \\[-1.8ex] 
0.00 (0.07) & 0.00 (0.07) & \{64\}/\{64, 64\} & \{64\}/\{64, 64\} & Peru & clay & Yes \\ 
0.37 (0.09) & 0.00 (0.09) & \{16\}/\{32, 64\} & \{16\}/\{16, 16\} & Uganda & clay & Yes \\ 
0.00 (0.09) & 0.11 (0.08) & \{349\}/\{349, 349\} & \{64\}/\{16, 64\} & Peru & clay & No \\ 
0.58 (0.10) & 0.01 (0.09) & \{16\}/\{32, 349\} & \{16\}/\{16, 32\} & Uganda & clay & No \\ 
0.68 (0.07) & 0.00 (0.06) & \{32\}/\{32, 64\} & \{64\}/\{64, 64\} & Peru & clip-rsicd & Yes \\ 
1.35 (0.10) & 0.19 (0.08) & \{64\}/\{32, 128\} & \{16\}/\{16, 32\} & Uganda & clip-rsicd & Yes \\ 
0.68 (0.09) & 0.00 (0.08) & \{32\}/\{32, 64\} & \{64\}/\{64, 64\} & Peru & clip-rsicd & No \\ 
0.95 (0.10) & 0.41 (0.09) & \{349\}/\{32, 128\} & \{16\}/\{16, 349\} & Uganda & clip-rsicd & No \\ 
0.06 (0.07) & 0.13 (0.06) & \{32\}/\{32, 64\} & \{16\}/\{16, 128\} & Peru & swin & Yes \\ 
0.09 (0.11) & 0.00 (0.10) & \{128\}/\{32, 128\} & \{128\}/\{128, 128\} & Uganda & swin & Yes \\ 
0.83 (0.09) & 0.29 (0.09) & \{32\}/\{32, 64\} & \{16\}/\{16, 32\} & Peru & swin & No \\ 
0.24 (0.11) & 0.00 (0.11) & \{64\}/\{16, 32\} & \{128\}/\{128, 128\} & Uganda & swin & No \\ 
\hline \\[-1.8ex] 
\end{tabular} 
}
\end{table}

\begin{table}[!htbp] \centering 
  \caption{Displaced RATE ratio differences from Equation \ref{eq:GlobalLoss}.
                                               Standard errors in parentheses.
                               ``clip-rsicd'' an EO fine-tune of CLIP.
                               PC denotes principal component representations.
                               $s^*, s_1^*, s_2^*$ denote optimal image 
                               dimensions in the single- and multi-scale cases using raw (uncompressed) representations. 
                               } 
  \label{tab:ClipDiffsDisplaced} 
\footnotesize 
\begin{tabular}{@{\extracolsep{5pt}} cccccc} 
\\[-1.8ex]\hline 
\hline \\[-1.8ex] 
PC: Multi-scale Gain & Multi-scale Gain & PC: \{$s^*$\}/\{$s_1^*$, $s_2^*$\} & \{$s^*$\}/\{$s_1^*$, $s_2^*$\} & Case & Model \\ 
\hline \\[-1.8ex] 
0.68 (0.07) & 0.00 (0.06) & \{32\}/\{32, 64\} & \{64\}/\{64, 64\} & Peru & clip-rsicd \\ 
1.35 (0.10) & 0.19 (0.08) & \{64\}/\{32, 128\} & \{16\}/\{16, 32\} & Uganda & clip-rsicd \\ 
\hline \\[-1.8ex] 
\end{tabular} 
\end{table}

  
\end{document}